\documentclass[sigconf,nonacm]{acmart}

\usepackage[utf8]{inputenc}
\usepackage[T1]{fontenc}
\usepackage{graphicx}
\usepackage{subcaption}
\usepackage[nolist]{acronym}
\usepackage{amsmath} % assumes amsmath package installed
\usepackage{array} % for \newcolumntype and \arraybackslash
\usepackage{booktabs} % for \toprule, \midrule, etc.
\begin{document}

\title{From Detection to Localization: A Unified Forensics Framework for
Fully Synthetic and Tampered Images}

\author{Annalisa Gallina}
%\orcid{0009-0003-6941-229X}
\correspondingauthor
\affiliation{%
  \institution{University of Padova}
  \city{Padova}
  \country{Italy}
}
\email{annalisa.gallina.1@phd.unipd.it}

\author{Marco Fiorucci}
\affiliation{%
  \institution{University of Padova}
  \city{Padova}
  \country{Italy}
}
\email{marco.fiorucci@unipd.it}
\author{Marco Brigo}
\affiliation{%
  \institution{University of Padova}
  \city{Padova}
  \country{Italy}
}
\email{marco.brigo@studenti.unipd.it}

\author{Federica Battisti}
\affiliation{%
  \institution{University of Padova}
  \city{Padova}
  \country{Italy}
}
\email{federica.battisti@unipd.it}
\author{Lamberto Ballan}
\affiliation{%
  \institution{University of Padova}
  \city{Padova}
  \country{Italy}
}
\email{lamberto.ballan@unipd.it}

\begin{abstract}
The rapid advancement of generative models has significantly worsened the problem of manipulated image detection, as these methods are capable of producing highly realistic forgeries, reinforcing the importance of multimedia forensics. Conventional approaches typically frame image manipulation detection as a binary classification task (real vs. generated), which limits the capability to distinguish and localize different forms of manipulation. To address these constraints, this work extends an existing detector by introducing a unified multiclass framework (real vs. fully generated vs. tampered).
In addition to classifying image authenticity, the framework incorporates a segmentation branch to enable pixel-level localization of tampered regions. The proposed approach outperforms selected recent benchmarks, offering an efficient solution with improved classification accuracy and higher IoU scores for the localization task. Find the code at \url{https://github.com/anngal01/From-Detection-to-Localization-A-Unified-Forensics-Framework-for-Fully-Synthetic-and-Tampered-Images.git}%\footnote{The code will be made available upon acceptance, not to reveal the authors’ identity.}.
%The code is available at \url{https://github.com/MarkBridge11/RINE-extended}.	
\end{abstract}

\maketitle

\begin{acronym}
\acro{GenAI}{Generative Artificial Intelligence}
\acro{GAN}{Generative Adversarial Network}
\acro{LVLM}{Large Vision-Language Model}
\acro{IFDL}{Image Forgery Detection and Localization}
\end{acronym}

\section{INTRODUCTION}
The rapid evolution of generative modeling has significantly changed the ways in which digital content is created and experienced.
Powered by large-scale architectures trained on massive open-world datasets, diffusion-based models have emerged as the prevailing approach for high-fidelity image synthesis~\cite{JonathanDenoising}, consistently surpassing adversarial methods~\cite{GoodfellowGan} in terms of quality and diversity. 
%such as  and latent diffusion frameworks~\cite{RombachDiffusion} %such as High-Resolution Image Synthesis with Latent Diffusion Models 
%have surpassed traditional \ac{GAN}-based methods~\cite{GoodfellowGan}, becoming the dominant paradigm for high-quality image generation. 
Current frameworks like latent diffusion~\cite{RombachDiffusion}, DALL·E 3~\cite{dalle3}, and Midjourney~\cite{ZexiColor}, enable the seamless generation of photorealistic visuals from textual prompts.
While these advancements have empowered creative workflow and augmented the digital content production, they simultaneously challenge the reliability of visual content~\cite{Pei2026ACMsurvey}. The ability to synthesize photorealistic imagery at scale facilitates the proliferation of misinformation and identity fraud, necessitating robust deepfake detection and \ac{IFDL} as the forensic focus shifts from obvious structural artifacts toward subtle, near-imperceptible statistical anomalies.
%This widespread accessibility has democratized content creation across digital art, design, marketing, journalism, and entertainment. However, the same technologies that empower creative expression also introduce serious societal risks. Highly realistic AI-generated images can be exploited for misinformation, political propaganda, identity manipulation, non-consensual explicit content generation, copyright infrinstegement, and financial fraud. Synthetic images may depict public figures in fabricated scenarios or fabricate events that never occurred, thereby distorting public perception and eroding trust in visual evidence. 
%As generative models continue to improve, 
%distinguishing synthetic images from authentic photographs becomes increasingly difficult, posing significant political, economic, legal, and ethical challenges.

To mitigate these emerging threats, the research community has intensified its efforts in the domain of visual forensics.
Early investigations primarily targeted specific architectural footprints and statistical inconsistencies unique to \ac{GAN}-generated content~\cite{ZhangGAN, ChaiGenrealization}. However, the transition toward diffusion-based synthesis has rendered many legacy techniques ineffective, as modern models often lack the upsampling artifacts and checkerboard patterns characteristic of previous generations. Consequently, recent literature has shifted toward exploring high-frequency inconsistencies, noise residuals, and reconstruction errors as robust forensic signatures~\cite{UtkarshUniversal, dire, YunpengLatent}. Furthermore, research has branched into spectral analysis, such as leveraging color distribution irregularities to enhance cross-domain robustness~\cite{ZexiColor}, and vision-language alignment, where prompt-tuned models harmonize visual and textual cues to detect semantic inconsistencies~\cite{Chang2023AntifakePromptPV}.

%while more recent approaches explore high-frequency inconsistencies, 
%noise patterns, reconstruction residuals, or semantic inconsistencies using large vision-language models ~\cite{UtkarshUniversal, ZhendongDIRE, YunpengLatent}. Other studies leverage color distribution irregularities to enhance detection robustness across domains ~\cite{ZexiColor}, or employ prompt-tuned vision-language models to align visual and textual cues for improved generalization ~\cite{Chang2023AntifakePromptPV}.
Despite these advancements, existing forensic frameworks remain hindered by critical technical bottlenecks. Most conventional detectors are formulated as closed-set binary classification tasks primarily designed to distinguish between entirely real and fully synthetic content~\cite{hifi, freaware}. While recent approaches such as RINE~\cite{ChristosRINE} and SPAI~\cite{karageorgiou2025any} have enhanced generalization across disparate architectures, by leveraging intermediate spectral invariants features, they remain predominantly centered on global image level labels.
This rigid formulation fails to address modern workflows where authentic images undergo generative tampering, such as diffusion-based inpainting, object insertion or scene-elements manipulation. These sophisticated manipulations necessitate a shift towards a multiclass paradigm capable of simultaneous detection and pixel level attribution, a transition recently facilitated by the introduction of new benchmarks \cite{ZhenglinSIDA, sofake}.
Additionally, while high-performing \acp{LVLM} offer forensic reasoning, their deployment is constrained by heavy computational overhead and a lack of pixel-level precision~\cite{ZhenglinSIDA, ZhipeiFakeShield}.

In this paper, we propose a unified framework that extends RINE~\cite{ChristosRINE} beyond binary classification. By identifying the "proxy token" phenomenon~\cite{wang2025declip} as a bottleneck in CLIP-based models, we adopt DINOv2~\cite{OquabDINOv2} to leverage superior spatial consistency through shared frozen representations. The architecture follows a sequential logic: a three-class branch categorizes images as real, fully synthetic, or partially tampered; the latter triggers a lightweight segmentation branch for pixel-level localization. This approach avoids extensive parameter tuning by exploiting the inherent discriminative power of fixed foundation features, significantly optimizing training efficiency. Extensive evaluations on SID-Set~\cite{ZhenglinSIDA} and So-Fake-Set~\cite{sofake} confirm that our framework matches or improves state-of-the-art performance, establishing a baseline for the dense forensic localization of diverse image content.

%The primary contributions of this work are: 1) we introduce a unified multiclass architecture for simultaneous detection and dense localization of real, synthetic, and tampered content; %2) we provide a technical ablation of Vision Foundation Models, proving DINOv2~\cite{OquabDINOv2} mitigates CLIP-based "proxy token" bottlenecks; and 
%2) we establish via extensive experiments on SID-Set~\cite{ZhenglinSIDA} and So-Fake-Set~\cite{sofake} that our framework matches or exceeds state-of-the-art performance with minimal training overhead.

%When an image is classified as tampered, a lightweight segmentation branch is activated, generating pixel-level localization masks that indicate manipulated regions. Our framework establishes a new state-of-the-art on the SIDA dataset~\cite{ZhenglinSIDA} while requiring substantially 
%fewer trainable parameters and training epochs. However, our generalization assessment reveals persistent challenges: performance degrades sharply on datasets containing manipulation styles absent from the training distribution, such as traditional Photoshop splices or early GAN artifacts. This domain shift sensitivity underscores the inherent difficulty of building a truly universal detector capable of identifying the full spectrum of possible image manipulations, from hand-crafted forgeries to modern AI-generated edits.
\begin{figure*}[t!]
\centering
\includegraphics[width=.9\textwidth]{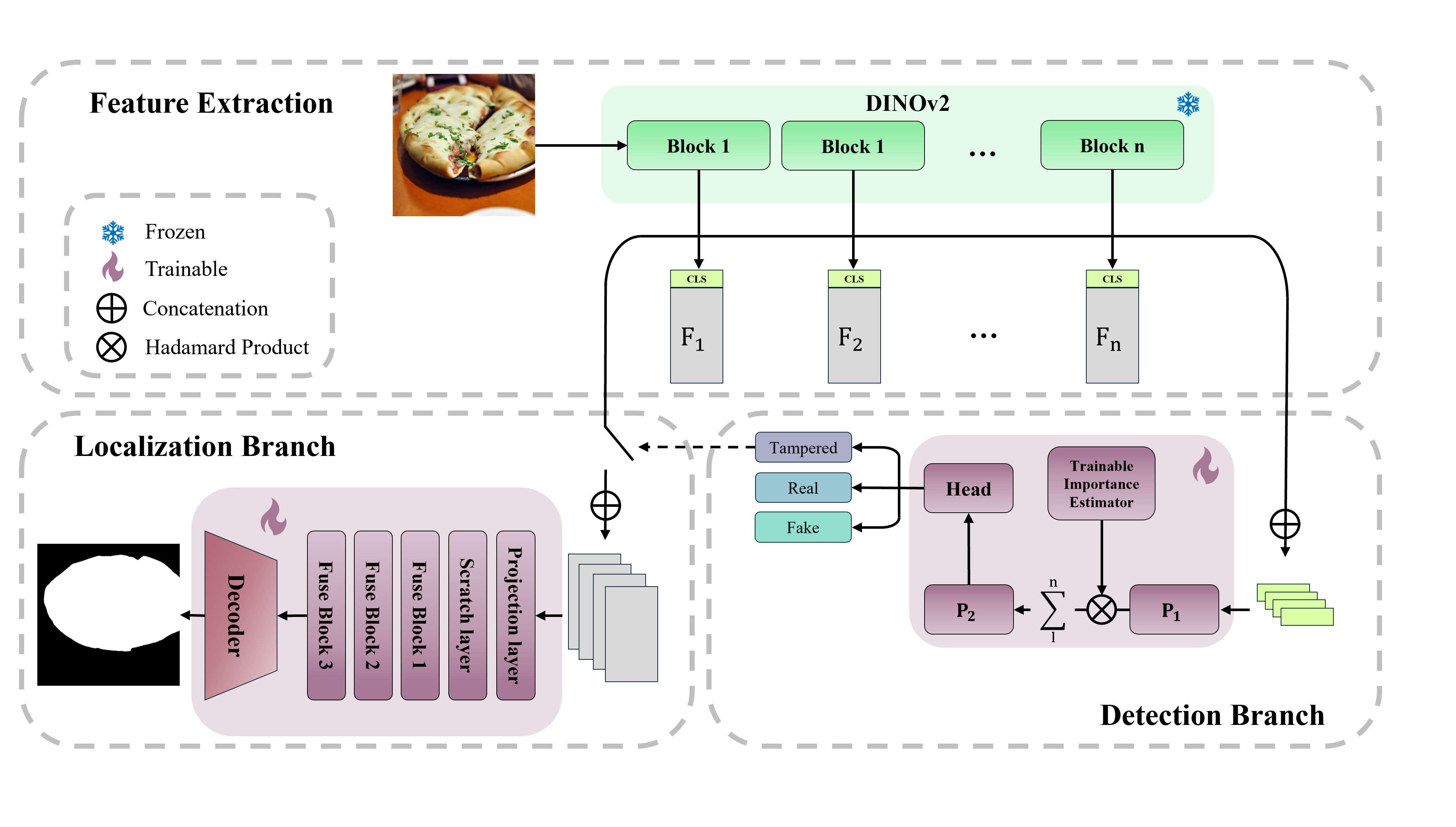}
\vspace{-2em} 
\caption{Overview of the proposed framework. An input image is processed by a frozen DINOv2 backbone, and features extracted from multiple transformer blocks are shared between two task-specific branches. The detection branch aggregates these representations through a trainable importance estimator to perform image-level classification, while the localization branch combines multi-scale features using a trainable decoder to predict the pixel-level tampering mask.}
\Description{Block diagram of the proposed framework. An input image is processed by a frozen DINOv2 backbone, and features F1 through Fn are extracted from multiple transformer blocks. The extracted features are shared by two branches. The classification branch projects and aggregates the features through a trainable importance estimator before a classification head predicts whether the image is real, fake, or tampered. The localization branch fuses multi-scale features through a decoder with three fuse blocks and a scratch layer to generate a binary tampering mask.}
\vspace{-1em} 
\label{pipeline}
\end{figure*}
\section{RELATED WORKS}
%Image deep-fake detection has become increasingly challenging  as generative models achieve unprecedented levels of photorealism. Early forensic studies focused on visible imperfections, such as misaligned features, inconsistent lighting, or unnatural textures~\cite{Li2018ExposingDV},~\cite{Matern2019ExploitingVA}. In modern image synthesis, however, these observable cues are largely mitigated, shifting the focus to imperceptible artifacts embedded within spatial and frequency domains~\cite{FreDect, Shiohara2022DetectingDW, Doloriel2024FrequencyMF}. 
%As a result, detection methods must extract imperceptible statistical irregularities that remain hidden to the human eye.
As generative models achieve unprecedented photorealism, deep-fake detection has shifted from visible imperfections~\cite{Li2018ExposingDV, Matern2019ExploitingVA} to imperceptible spatial and frequency artifacts~\cite{FreDect, Shiohara2022DetectingDW, Doloriel2024FrequencyMF}. Consequently, modern methods must extract subtle statistical irregularities hidden from the human eye.\\

\textbf{Spatial and Frequency Artifacts}.

In the spatial domain, pixel-level irregularities, including boundary blurring, blending inconsistencies, can be captured by training detectors on self-blended images~\cite{Shiohara2022DetectingDW}.  
%demonstrated that models can learn robust forgery patterns by training on self-blended images, thereby reducing the reliance on generator-specific fingerprints. 
Nevertheless, as generative architectures achieve higher fidelity and resolution, these spatial cues become increasingly subtle and are often attenuated by standard post-processing operations, such as lossy compression, rescaling, or enhancement filters. Consequently, detectors often exhibit a significant performance degradation when evaluated across diverse datasets or previously unseen generative paradigms~\cite{yang2025d3}.

%Frequency-domain analysis provides a complementary perspective by isolating artifacts induced by convolutional and upsampling operations. 
Frequency domain analysis offers a complementary perspective by isolating upsampling-induced spectral anomalies\cite{FreDect, Doloriel2024FrequencyMF}, yet these signatures remain strongly architecture-dependent.   
%These spectral signatures, often imperceptible in the spatial domain, reveal the underlying mathematical regularities of generative models.
%These operations often manifest as characteristic spectral peaks or anomalous energy distributions in the Fourier domain~\cite{FreDect, Doloriel2024FrequencyMF}. 
While GAN-based models, such as ProGAN~\cite{Karras2017Progressive} and StyleGAN2~\cite{sevastopolsky2023boost} exhibit prominent spectral spikes~\cite{Corvi2022OnTD}, diffusion-based frameworks, like Latent Diffusion~\cite{JonathanDenoising}, Stable Diffusion~\cite{RombachDiffusion}, DALL·E, and Midjourney~\cite{ZexiColor} exhibit attenuated or absent peaks.
 %Specifically, although Latent Diffusion~\cite{JonathanDenoising} and Stable Diffusion~\cite{RombachDiffusion} retain detectable frequency patterns, novel diffusion models such as DALL·E and Midjourney~\cite{ZexiColor} exhibit significantly weaker or non-existent peaks. 
Consequently, this spectral heterogeneity, paired with low perturbation robustness, limits the transferability of frequency-only forensic frameworks.\\
%This spectral heterogeneity suggests that frequency artifacts are often architecture-dependent rather than universal indicators of synthetic content. Furthermore, these spectral signatures are highly susceptible to degradation, mirroring the fragility of spatial artifacts. Their lack of robustness against routine operations and their diminished transferability to unseen generators underscore the limitations of frequency-only forensic frameworks.

\textbf{Image Deepfake Detection and Localization}.

To address the inherent challenges posed by subtle and generator-dependent artifacts, many deepfake detection methods employ data-driven classification paradigms, leveraging CNNs or Vision Transformers (ViTs). 
Many frameworks adopt multimodal fusion strategies, combining spatial textures and frequency-domain descriptors to enhance discriminative power~\cite{hifi, freaware}. 
%Rather than relying on manually defined forensic cues, these approaches aim to learn robust representations by leveraging diverse backbones, ranging from Convolutional Neural Networks (CNNs) to Vision Transformers (ViTs). To mitigate the impact of signal degradation, many current frameworks implement multimodal fusion strategies, effectively integrating spatial textures with frequency-domain descriptors to enhance their discriminative power~\cite{hifi, freaware}.
While models trained on ProGAN~\cite{Gao2019ProGAN} may retain intrafamily consistency on related architectures like StyleGAN~\cite{Sauer2023StyleGAN}, they often suffer from performance collapse when encountering fundamentally different frameworks like Diffusion Models~\cite{yang2025d3, dire}.
%Limitatiopn binary classification
%Detectors often overfit to generator-specific patterns, performing well on familiar GANs but struggling with diffusion-based models or unseen architectures~\cite{yang2025d3}. 
%However, a critical limitation of current methodologies is still their tendency to overfit to generator-specific fingerprints, leading to a substantial generalization gap when evaluated on images produced by unseen or different generative models~\cite{yang2025d3}.
%This phenomenon is particularly evident when evaluating models across architecturally related families; for example, a framework trained on ProGAN~\cite{Gao2019ProGAN} may maintain a degree of intrafamily consistency against related variants like StyleGAN~\cite{Sauer2023StyleGAN} due to shared upsampling behaviors.
%However, such detectors frequently suffer from performance collapse when encountering fundamentally different synthesis paradigms, such as Diffusion Models~\cite{yang2025d3, dire}. This suggests current forensic cues are tied to specific architectural biases rather than universal features of synthesis.
To mitigate these discrepancies, recent works have explored intermediate feature representations to improve robustness across unseen generators. Specifically, RINE~\cite{ChristosRINE} demonstrates that representations extracted from intermediate encoder blocks capture more transferable forensic information than final-layer features, which often suffer from generator-specific bias. By leveraging these multi-level representations, the method enhances robustness across both GAN- and diffusion-based synthesis.
Similarly, SPAI~\cite{karageorgiou2025any} addresses the generalization challenge by adopting the spectral distribution of real images as a universal reference signal. This framework employs a Vision Transformer to reconstruct masked frequency components using exclusively real images. At test time, samples with high reconstruction errors are flagged as out-of-distribution and classified as synthetic, improving detector robustness against novel generation models.

Despite these advances, most deepfake detection approaches remain binary, either distinguishing fully synthetic images from real ones or detecting whether specific regions have been tampered. Even in localization tasks, where pixel-level masks of manipulated areas are available, existing models operate within a two-class framework centered solely on the presence of manipulation~\cite{pscc,TruFor}.
%Multiclass
A primary bottleneck in moving beyond binary classification has been the scarcity of datasets designed for unified multi-class settings (i.e., encompassing real, fully synthetic, and locally tampered images simultaneously). Existing benchmarks typically isolate synthetic detection from forgery localization, hindering the development of models capable of handling heterogeneous manipulations. Recently, benchmarks such as SID-Set~\cite{ZhenglinSIDA} and So-Fake Set~\cite{sofake} have begun to address this limitation by offering larger, more diverse collections for comprehensive forensic evaluation.

To bridge this gap, we propose a multi-class deepfake detection framework that jointly distinguishes between real, synthetic, and tampered images. Unlike traditional approaches that isolate global generation from local forgery, our architecture classifies these diverse manipulation types within a unified formulation, overcoming the structural constraints of conventional binary detectors. \\

\textbf{Large Vision-Language Models}.

Recent advances in \ac{LVLM} have motivated their application to image deepfake detection and localization tasks. Unlike specialized binary classifiers, \ac{LVLM}s combine visual perception with linguistic reasoning, enabling a more nuanced analysis of manipulated content. General purpose models, such as~\cite{llava, deepseek, bai2025qwen25, internvl3}, demonstrate robust visual reasoning capabilities and can be adapted for the \ac{IFDL} task through strategic prompting or supervised fine tuning.

Beyond general architectures, several works introduce task-specific extensions. For instance, LISA~\cite{lisa} integrates a dedicated segmentation decoder for fine-grained localization, while SIDA~\cite{ZhenglinSIDA} proposes a unified multimodal framework for deepfake detection, localization, and explanation on social media imagery. FakeShield~\cite{ZhipeiFakeShield} builds upon LLaVA to localize manipulated regions while generating interpretable rationale, and So-Fake-R1~\cite{sofake} fine-tunes an \ac{LVLM} via reinforcement learning to enhance forensic reasoning. Yet, despite their promising generalization and reasoning abilities, these approaches suffer from high computational complexity, limiting their deployment in real-world, latency-sensitive applications.

To address these latency and resource constraints, we move beyond heavy multimodal models in favor of a unified, highly efficient framework. Our architecture jointly handles image-level classification and pixel-level localization, delivering competitive forensic precision at a fraction of the computational cost.

\section{PROPOSED METHOD}
%\subsection{Overall Architecture}
%As summarized in Figure \ref{pipeline}, the proposed framework consists of a two-branch architecture for joint image-level detection and pixel-level localization, built upon a shared frozen DINOv2 backbone.
%Given an input image, the backbone extracts transformer representations that are reused by both branches. The Detection Branch operates at the image level and produces a three-class prediction (Real, Tampered, Fake) while the Localization Branch, which is conditionally activated only when an image is classified as Tampered, exploits multi-scale patch tokens from intermediate transformer layers to generate a binary segmentation mask.
%By sharing the same frozen backbone, the architecture ensures computational efficiency while jointly modeling global semantic inconsistencies and fine-grained spatial artifacts. In the following, the main blocks will be described.
As summarized in Figure~\ref{pipeline}, the proposed framework consists of a two-branch architecture for joint image-level detection and pixel-level localization of manipulated areas, built upon a shared DINOv2 backbone. Using a common set of frozen representations, the model maintains computational efficiency while jointly capturing global semantic inconsistencies and fine-grained artifacts. The main architectural components are described in the following sections.

\subsection{Detection Branch}
The Detection Branch performs image-level classification and constitutes the first stage of the framework. Its objective is to assign each input image to one of three classes -- Real, Tampered, or Fake -- based on the representations extracted by the frozen backbone.\\

\textbf{Feature Extraction}
With the rise of large foundation models, features extracted from these networks have become effective content descriptors. In this context, prior work has shown that combining high-level semantic features with low-level image cues improves performance in image forgery detection~\cite{ChristosRINE}. However, not all pre-trained models are equally suitable for this task. Models that do not capture fine-grained image content may reduce the ability to detect manipulations, which makes the selection of the backbone crucial for \ac{IFDL} performance.

As argued in~\cite{wang2025declip}, CLIP-based models are trained for image-level understanding and often struggle with dense, region-level tasks such as tampering localization. Their attention maps tend to focus on a limited set of "proxy" tokens, overlooking subtle local inconsistencies. As a result, CLIP features lack local discriminability and spatial consistency, which limits their effectiveness in detecting small artifacts.

In contrast, DINOv2~\cite{OquabDINOv2} does not exhibit this behavior. Its attention maps remain aligned with the image content, so that when the reference patch changes position, the model continues to focus on semantically related regions. This results in more stable features, with stronger inter-patch correlations and more consistent representations across the image. For these reasons, DINOv2 is a more suitable choice as a frozen backbone. 

Specifically, following~\cite{ChristosRINE}, a batch $b$ of input images is processed by the encoder, producing $n$ CLS tokens of dimension $d$. These tokens are concatenated and passed through a learnable projection network ($P_1$ in Figure~\ref{pipeline}). To account for the different relevance of the extracted features, a Trainable Importance Estimator (TIE) module is used to assign weights to each token. Summation over the second dimension then yields a single feature vector per image, which is further projected by a second network ($P_2$ in Figure~\ref{pipeline}) with the same architecture as $P_1$. The resulting representation is finally fed into the classification head.\\

\textbf{Classification Head}
Given the extracted feature vector $f \in \mathbb{R}^d$, the classification head computes the final detection probabilities. Unlike most previous approaches, which formulate the problem as binary classification, we address a three-class setting distinguishing among Real, Tampered, and Fake images.

To this end, we employ a lightweight MLP composed of two fully connected layers of size $d \times d$, each followed by a ReLU activation, and a final linear layer of size $d \times 3$. The network outputs a three-dimensional logit vector, which is converted into class probabilities through a softmax function.\\

\textbf{Training Objective}
Inspired by~\cite{ChristosRINE}, the loss function used to optimize the detector parameters consists of two components: a Categorical Cross-Entropy loss, denoted as $\mathcal{L}_{CCE}$, and a Supervised Contrastive Loss, denoted as $\mathcal{L}_{Cont}$ ~\cite{supervised}.

The Categorical Cross-Entropy loss is a standard objective for multi-class classification tasks. It measures the discrepancy between the predicted class probabilities and the ground-truth distribution, thereby directly optimizing the \ac{IFDL} task. It is defined as:

\begin{equation*} \mathcal{L}_{CCE} = - \sum_{i=1}^C y_i \log (\hat{y}_i) \end{equation*}
where $y_i$ denotes the ground-truth probability for class $i$, $\hat{y}_i$ the predicted probability for class $i$, and $C$ the number of classes.

The Supervised Contrastive Loss~\cite{supervised} is introduced to improve the discriminative structure of the learned feature space. Specifically, it encourages feature vectors associated with the same class label to be closer, while pushing apart those corresponding to different classes. It is defined as:
\begin{equation*} \mathcal{L}_{Cont} = -\sum_{i=1}^{b} \frac{1}{G(i)} \sum_{g \in G(i)} \log \frac{\exp(\mathbf{z}_i \cdot \mathbf{z}_g / \tau)}{\sum_{a \in A(i)} \exp(\mathbf{z}_i \cdot \mathbf{z}_a / \tau)} \end{equation*} where $A(i) = \{1, \ldots, i-1, i+1, \ldots, b\}, \quad G(i) = \{g \in A(i) : y_g = y_i\}
%, \quad z_j = \left\{\tilde{\mathbf{K}}_{q}^{(jk)}\right\}_{k=1}^{d'} \in \mathbb{R}^{d'}
$, $\cdot$ denotes the dot product and $\tau$ is the temperature parameter.

The overall loss is defined as a linear combination of the two components:
\begin{equation*} \mathcal{L} = \mathcal{L}_{CCE} + \xi \mathcal{L}_{Cont} \end{equation*}
where $\xi$ is a tunable hyperparameter controlling the relative contribution of the contrastive term. This combined objective promotes both accurate classification and well-structured latent representations, which in turn enhances the overall model performance.

\subsection{Localization Branch}
The Localization Branch is designed to generate pixel-level binary masks that identify manipulated regions within input images. As illustrated in the bottom-left corner of Figure~\ref{pipeline}, this module operates conditionally: it is activated only when the image is classified as tampered by the preceding stage. When triggered, the branch leverages the representations learned by the frozen backbone and is organized into three sequential stages: feature extraction, progressive feature fusion, and final mask decoding.\\

\textbf{Feature Extraction}
Multi-scale transformer features are extracted from the frozen backbone to support dense pixel-level prediction. To maintain a lightweight design, we reuse computations from the classification forward pass. Specifically, while CLS tokens are extracted for image-level classification, the patch tokens, commonly discarded in prior approaches, are retained and further processed for localization.

Previous studies~\cite{ranftl2021vision} have shown that combining patch tokens from intermediate transformer layers significantly benefits dense prediction tasks. Such aggregation enables the joint exploitation of low-level structural information and high-level semantic representations. Accordingly, we extract features from layers $\{3,6,9,12\}$ of the DINOv2 base model~\cite{OquabDINOv2}. Since DINOv2 keeps the same patch resolution across all layers, the fine spatial details captured in early representations remain aligned and preserved throughout the network. Because of this, the fine-grained spatial information captured in early layers is not lost as features become more abstract. At the same time, deeper layers still learn higher-level semantic representations, making the features well suited for segmentation.

To recover the spatial structure, the output sequence of token embeddings is reshaped into 2D feature maps that reflect the layout of the original input patches. Each map is then processed with a $1 \times 1$ convolution to project it into a higher-dimensional embedding space, increasing channel capacity and enhancing representational power. Finally, scratch layers are applied to standardize the channel dimension across feature maps, ensuring compatibility before their subsequent fusion.\\

\textbf{Feature Fusion}
\begin{figure}[t]
    \centering
    \begin{subfigure}[b]{0.47\linewidth}
        \centering
        \includegraphics[width=\textwidth]{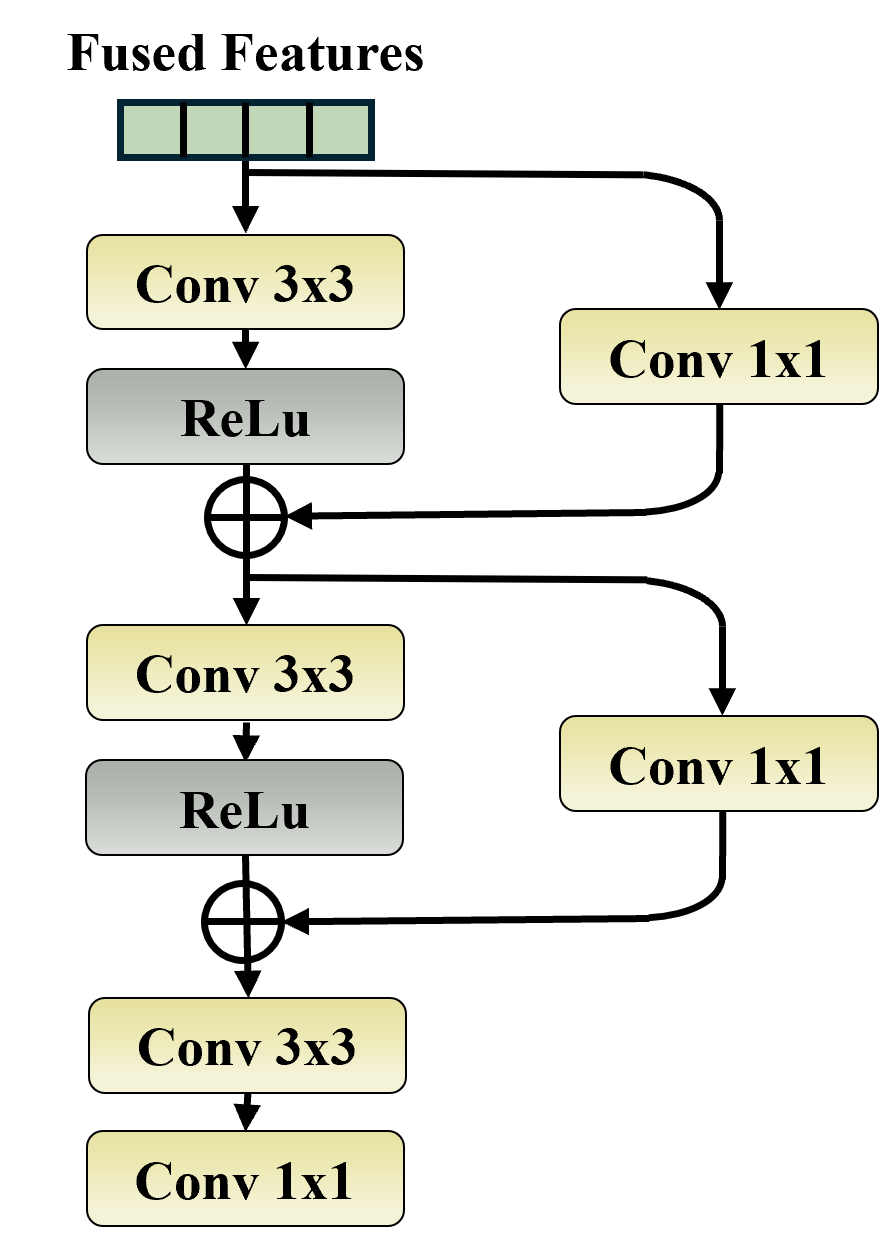}
        \caption{}
        \label{fig:decoder}
    \end{subfigure}
    \hfill
    \begin{subfigure}[b]{0.505\linewidth}
        \centering
        \includegraphics[width=\textwidth]{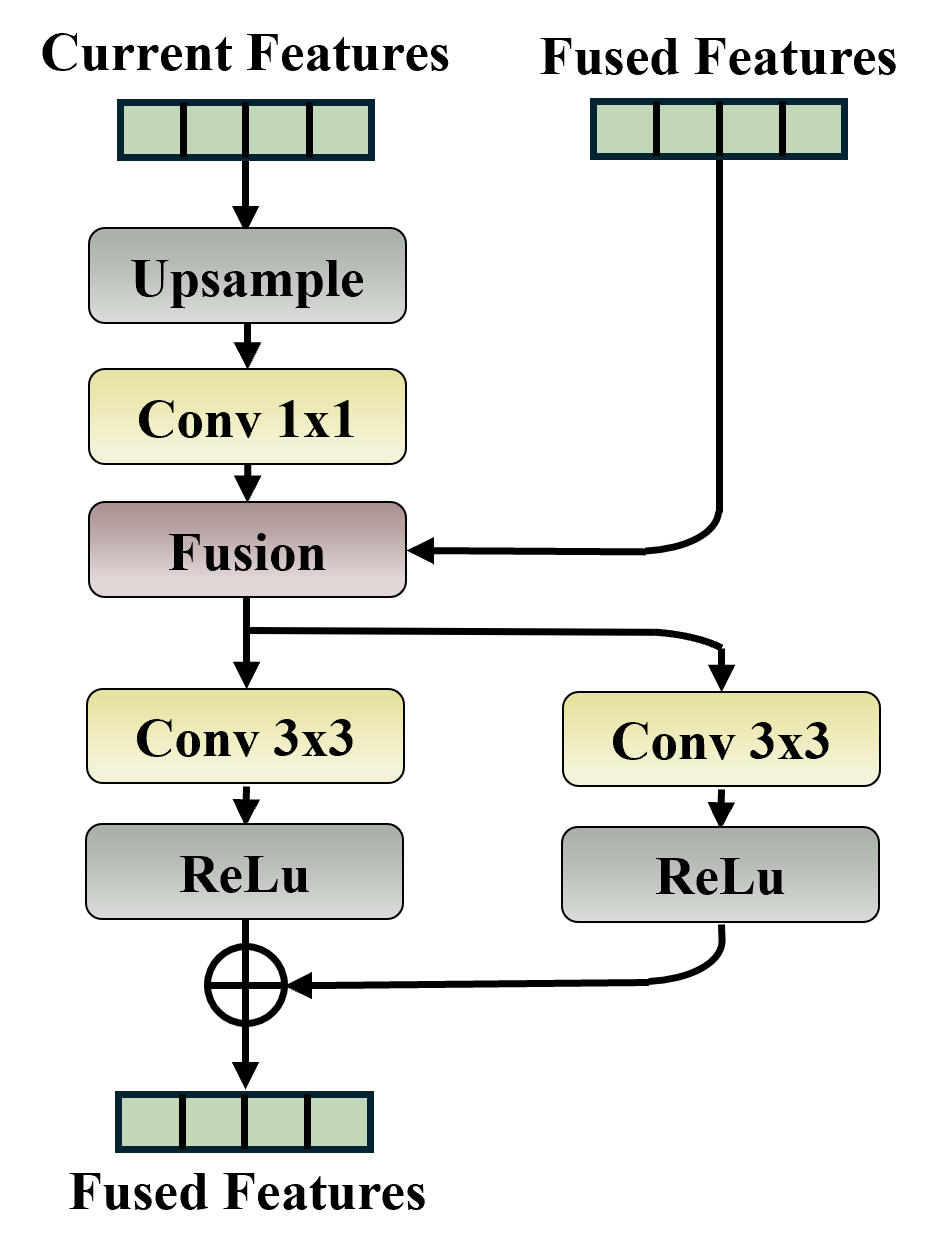}
        \caption{}
        \label{fig:fusion_block}
    \end{subfigure}
    \vspace{-1em}
    \caption{Illustration of the decoder architecture (a) and the fusion block (b).}
    \Description{Two flowcharts illustrating the feature fusion modules. In (a), a residual refinement block processes fused features through a sequence of convolutional layers with ReLU activation and additive skip connections, producing refined fused features. In (b), a fusion block combines current and previously fused features through upsampling, a $1 \times 1$ convolution, and a fusion operation, followed by parallel convolutional paths with ReLU activation and an additive connection to generate the output fused features.}
    \label{fig:comparison}
\end{figure}
The features extracted from different backbone layers encode complementary information, ranging from low-level details to high-level semantic representations. To effectively exploit this hierarchy, we adopt a progressive fusion strategy in which deeper features are gradually upsampled and integrated with shallower feature maps. 

The proposed fusion block is illustrated in Figure~\ref{fig:fusion_block}: before fusion, each feature map is passed through a $1 \times 1$ convolutional alignment layer to project it into a shared embedding space, ensuring channel-wise semantic consistency. In line with~\cite{ranftl2021vision}, aggregation is performed through element-wise addition rather than concatenation, which maintains dimensional efficiency and avoids introducing additional parameters. The fused features are subsequently refined using a two-stage residual refinement module, enhancing the quality of the final representation. 

Since four backbone layers are considered, three successive fusion stages are performed, yielding the final fused feature representation.\\

\textbf{Mask Decoding}
A U-Net–style decoder processes the fused feature representation to produce the final binary segmentation mask, as illustrated in Figure~\ref{fig:decoder}.

The decoder consists of two upsampling stages, each incorporating residual connections to preserve information flow. In the first stage, channel dimensionality is reduced from 256 to 128 via a $1 \times 1$  convolution followed by GroupNorm~\cite{groupnorm} and ReLU, while a parallel $1 \times 1$ convolution provides a residual shortcut on the upsampled input. A second equivalent stage further compresses the representation to 64 channels to limit information loss. Following these upsampling steps, an additional $3 \times 3$ refinement block enhances spatial definition before a final $1 \times 1$ convolution that maps the refined features to a single-channel output corresponding to the binary segmentation mask.\\
\begin{table*}[t]
    \centering
    \newcolumntype{C}{>{\centering\arraybackslash}p{10em}}
    \newcolumntype{K}{>{\centering\arraybackslash}p{6em}}

    \begin{tabular}{C *{6}{K}}
        \toprule
        Method  & Year & Type &  \multicolumn{2}{c}{Detection} & \multicolumn{2}{c}{Localization} \\
        \cmidrule(lr){4-5} \cmidrule(lr){6-7} 
           & &&Accuracy & F1 & IoU & F1 \\
        \midrule
        CnnSpot~\cite{cnnspot}& 2021 & Detection & 89.6 & 87.7 & - & - \\
        UnivFD~\cite{UtkarshUniversal} & 2023 & Detection & 84.0 & 63.8 & - & -  \\
        FreAware~\cite{freaware} & 2024 & Detection & 85.6 & 73.1 & - & - \\
        NPR~\cite{NPR} & 2024 & Detection & 81.8 & 61.5 & - & - \\
        HIFI-Net~\cite{hifinet} & 2022 & IFDL & 39.0 & 25.2 & 12.1 & 18.3 \\
        TruFor~\cite{TruFor} & 2023 & IFDL & 87.3 & 85.9 & 47.5 & 57.6 \\
        PSCC-Net~\cite{pscc}& 2022 & IFDL & 84.2 & 81.1 & 46.3 & 54.8 \\
        FakeShield~\cite{ZhipeiFakeShield} & 2024 & LLM & 67.0 & 64.1 & 33.7 & 46.1 \\
        SIDA~\cite{ZhenglinSIDA}& 2025 & LLM & 91.9 & 91.5 & 44.1 & 58.9 \\
        LLaVA-1.5-13B~\cite{llava}& 2024 & LLM & 83.5 & 82.9 & 29.8 & 38.1 \\
        LISA~\cite{lisa} & 2024 & LLM & 87.4 & 85.9 & 40.5 & 47.6 \\
        DeepSeek-VL-7B~\cite{deepseek}& 2025 & LLM & 83.7 & 81.1 & 27.8 & 35.4 \\
        Qwen2.5-VL-7B~\cite{bai2025qwen25}& 2024 & LLM & 91.2 & 90.0 & 42.7 & 50.1 \\
        InternVL3-8B~\cite{internvl3}& 2025 & LLM & 87.6 & 87.3 & 41.1 & 48.5 \\
        So-Fake-R1~\cite{sofake} & 2025 & LLM & \textbf{93.2} & \textbf{92.9} & \underline{48.6} & \underline{63.9} \\
        \midrule
        Ours & 2026 & IFDL & \underline{92.4} & \underline{92.4} & \textbf{77.8} & \textbf{83.9}\\
        \bottomrule
    \end{tabular}
    \caption{Performance comparison on So-Fake-Set: best results are highlighted in \textbf{bold} while second best are \underline{underlined}.}    
    \label{tab:results}
    \vspace{-1em}
\end{table*}

\textbf{Training Objective}
Following ~\cite{ZhenglinSIDA}, the adopted loss function to train the Segmentation Branch is a weighted combination of the Dice loss, denoted as $\mathcal{L}_{DICE}$, and the Binary Cross-Entropy (BCE) loss, denoted as $\mathcal{L}_{BCE}$.

The BCE loss is defined as:
\begin{equation*} \mathcal{L}_{BCE} = -\frac{1}{n} \sum_{i=1}^{n} \left( Y_i \log \hat{Y}_i + (1 - Y_i) \log (1 - \hat{Y}_i) \right) \end{equation*}
where $\hat{Y}_i$ denotes the predicted probability for pixel $i$, $Y_i$ the corresponding ground-truth label, and $n$ the total number of pixels. It is a standard objective for binary classification and segmentation tasks but, in the presence of severe class imbalance, it may bias the optimization toward the majority class.

Since manipulated pixels typically constitute a minority within an image, we incorporate the Dice loss, which has been shown to be effective for highly imbalanced segmentation problems~\cite{dice}. It is defined as:
\begin{equation*}
\mathcal{L}_{\mathrm{DICE}} = 1 - \frac{2|X \cap Y| + \epsilon}{|X| + |Y| + \epsilon}
\end{equation*}
where $X$ represents the predicted mask, $Y$ the ground-truth mask, and $\epsilon$ a small constant introduced for numerical stability. The Dice loss directly maximizes the overlap between predicted and reference regions, making it particularly sensitive to spatial alignment and region-level consistency.

The overall objective function is given by:
\begin{equation*} \mathcal{L} = \mathcal{L}_{BCE} + \mathcal{L}_{DICE}. \end{equation*}
By combining these two terms with equal weights, the model jointly enforces accurate pixel-wise classification through BCE and robust region-level agreement through Dice, resulting in more precise and spatially coherent mask predictions.

\section{EXPERIMENTAL RESULTS}

\subsection{Experimental Setting}
\textbf{Implementation details}
The proposed system is trained and evaluated on the So-Fake dataset~\cite{sofake}, which represents, to the best of our knowledge, the most comprehensive non-binary benchmark for fake image detection and localization. To ensure a fair comparison, we follow the training and test splits defined in the original study.

The architecture is optimized separately for the classification and segmentation tasks. The classification branch is trained on the entire training set, while the segmentation branch is trained exclusively on tampered images, since localizing manipulated regions is not meaningful for pristine or fully synthetic samples.

For the multi-class classification branch, we adopt the ADAM optimizer with a learning rate of $1 \times 10^{-3}$ and a batch size of 128. Training is limited to three epochs, as additional epochs yield limited performance gains. To identify the optimal model configuration, we perform a grid search over the hyperparameters $\xi \in \{0.2, 0.4\}$ and $d \in \{512, 1024\}$ while fixing the number of projections to 2. The best-performing configuration is obtained with $\xi = 0.2$ and $d = 512$. Unless otherwise specified, feature extraction is performed using DINOv2 pretrained on the LVD-142M dataset~\cite{OquabDINOv2} as the backbone network.

For the segmentation branch, training is conducted for 5 epochs using a batch size of 128 and a learning rate of $1 \times 10^{-3}$, with ADAM as the optimizer.
%All experiments, including both training and evaluation, are conducted on a NVIDIA RTX A6000 GPU.
\\

\textbf{Preprocessing}
To comply with the input size requirements of DINOv2 ~\cite{OquabDINOv2}, images are center-cropped to $518 \times 518$. In addition, standard data augmentation techniques are applied during training to improve generalization. Gaussian blur and JPEG compression are applied with probability $p = 0.5$, followed by random horizontal flipping with probability $p = 0.5$. %Distortion-based augmentations and flipping are used for both classification and segmentation, whereas random cropping is applied only to the classification branch.

\subsection{Baselines and Evaluation Metrics}

We evaluate the performance of the proposed method against several baselines, including CNN-Spot~\cite{cnnspot}, UnivFD~\cite{UtkarshUniversal}, FreeAware~\cite{freaware}, NPR~\cite{NPR}, HIFI-Net~\cite{hifinet}, TruFor~\cite{TruFor}, PSCC-Net~\cite{pscc}, FakeShield~\cite{ZhipeiFakeShield}, and SIDA~\cite{ZhenglinSIDA}. In addition, we consider several \ac{LVLM}s, namely LLaVA-1.5-13B~\cite{llava}, LISA~\cite{lisa}, DeepSeek-VL-7B~\cite{deepseek}, Qwen2.5-VL-7B~\cite{bai2025qwen25}, InternVL3-8B~\cite{internvl3}, and So-Fake-R1~\cite{sofake}. All methods are evaluated on the detection task, while only LLM-based approaches and methods that explicitly support localization are considered for the localization task.

For classification, performance is measured in terms of image-level accuracy and F1 score. For segmentation, performance is assessed using Intersection over Union (IoU) and F1 score.

Unless otherwise specified, results for the baselines are taken from~\cite{sofake}, where most models are fine-tuned on the So-Fake dataset to ensure a fair comparison. Two exceptions are considered: FakeShield~\cite{ZhipeiFakeShield}, which requires paired image–text inputs and is evaluated using its publicly available checkpoints, and HIFI-Net~\cite{hifinet}, which is evaluated using pre-trained weights due to the unavailability of the complete training code.

\subsection{Quantitative Results}

Table~\ref{tab:results} presents the comparison with state-of-the-art methods on the So-Fake validation set. The proposed method achieves competitive detection performance while significantly improving localization accuracy. Specifically, it obtains an accuracy of $92.4\%$ on the first task, closely matching the top-performing method (So-Fake-R1) and outperforming all the other existing baselines. For the localization task, we evaluate the predicted masks on all tampered images in the validation set, regardless of whether the corresponding image is correctly identified as tampered by the detection branch to decouple localization from detection performance.Under this setting, our method achieves an IoU of $77.8\%$, surpassing previous methods by a large margin and demonstrating an improved ability to precisely localize manipulated regions.
\\
% Requires: \usepackage{booktabs}
\begin{table}[h]
    \centering
    \begin{tabular}{l cc cc cc}
        \toprule
        Methods & \multicolumn{2}{c}{Real} & \multicolumn{2}{c}{Fake} & \multicolumn{2}{c}{Overall} \\
        \cmidrule(r){2-3} \cmidrule(r){4-5} \cmidrule(r){6-7}
        & Acc & F1 & Acc & F1 & Acc & F1 \\
        \midrule
        CnnSpot      & 89.0 & 90.8 & 79.4 & 76.1 & 84.2 & 83.5 \\
        Gram-Net      & 89.2 & 91.7 & 93.9 & 92.8 & 91.6 & 92.3 \\
        Fusing        & 89.2 & 92.7 & 57.6 & 60.3 & 73.4 & 76.5 \\
        UnivFD        & 68.3 & 68.5 & 89.5 & 94.0 & 78.9 & 81.3 \\
        AntifakePrompt & 88.9 & 89.1 & 94.2 & 89.2 & 91.6 & 89.2 \\
        SIDA-7B       & 89.1 & 91.0 & 95.0 & 94.8 & 92.1 & 92.9 \\
        So-Fake-R1    & 91.1 & 92.9 & 95.6 & 95.1 & 93.4 & 94.0 \\
        Ours          & \underline{96.0} & \underline{96.6} & \underline{99.7} & \underline{99.7} & \underline{97.7} & \underline{97.7} \\
        \bottomrule
    \end{tabular}
    \caption{Performance comparison on SID-Set: best results are highlighted \underline{underlined}.}
    \label{tab:crossdataset}
    \vspace{-2em}
\end{table}

\textbf{Cross-dataset evaluation}
To evaluate the generalization capability of the proposed approach, we evaluated the model trained on So-Fake~\cite{sofake} on SID-Set~\cite{ZhenglinSIDA}, without any further training or fine-tuning on the target dataset. Following the protocol in~\cite{sofake}, we compare with CnnSpot~\cite{cnnspot}, Gram-Net~\cite{gramnet}, Fusing~\cite{fusing}, UnivFD~\cite{UtkarshUniversal}, AntifakePrompt~\cite{Chang2023AntifakePromptPV}, SIDA~\cite{ZhenglinSIDA}, and So-Fake-R1~\cite{sofake}. For consistency with~\cite{sofake}, we report results only for the Real and Fake classes, without modifying the model's three-class output. As shown in Table~\ref{tab:crossdataset}, the proposed method achieves the best overall performance, indicating strong cross-dataset generalization. The authors in~\cite{sofake} do not report cross-dataset localization results; nevertheless, our method achieves good performance with an IoU of 72.95 on this setting, further supporting its generalization capabilities.

\subsection{Qualitative Results}
\begin{figure*}[t!]
\centering

\setlength{\tabcolsep}{2pt} % spazio tra colonne

\begin{tabular}{c c c c c c c}
& \textbf{(a)} & \textbf{(b)} & \textbf{(c)} & \textbf{(d)} &\textbf{(e)} & \textbf{(f)}\\

\rotatebox{90}{ $\quad$ FLUX.1-dev} &
\includegraphics[width=0.133\linewidth]{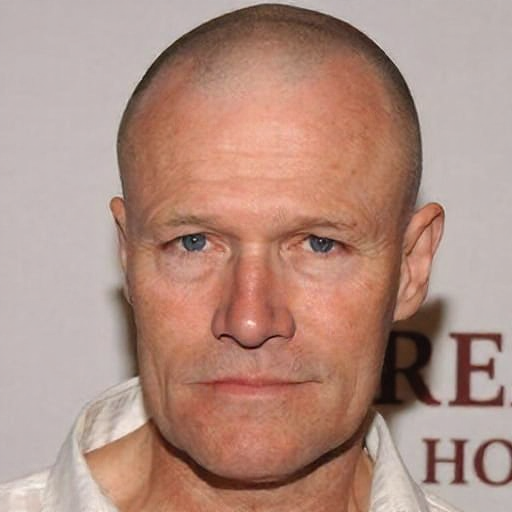} &
\includegraphics[width=0.133\linewidth]{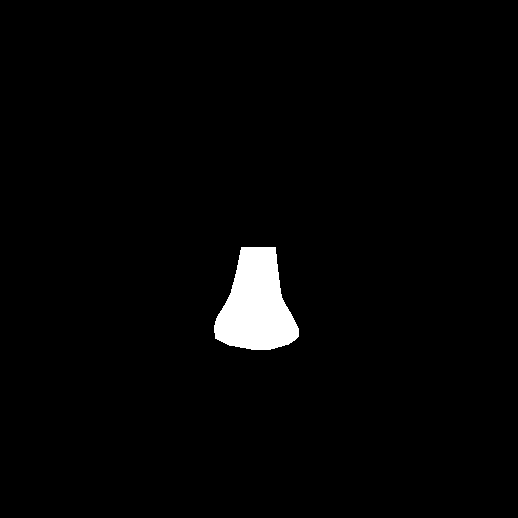} &
\includegraphics[width=0.133\linewidth]{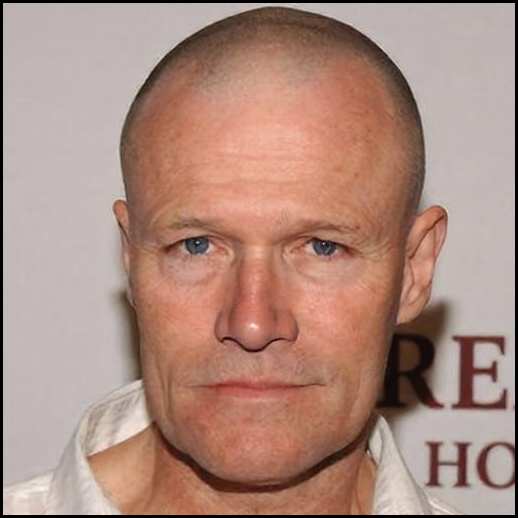} &
\includegraphics[width=0.133\linewidth]{29007_generated_34_iou_0.935_gt_mask.png} &
\includegraphics[width=0.133\linewidth]{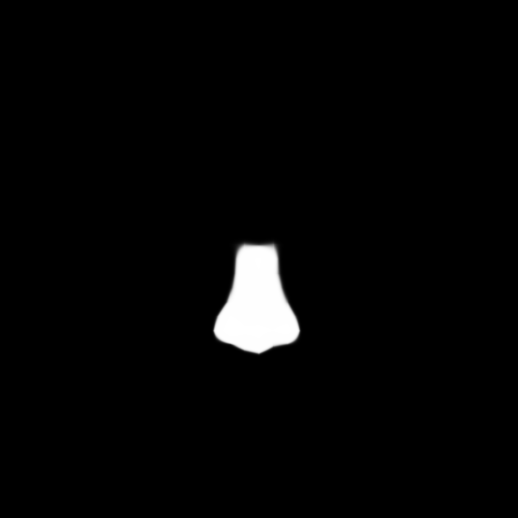}&
\includegraphics[width=0.133\linewidth]{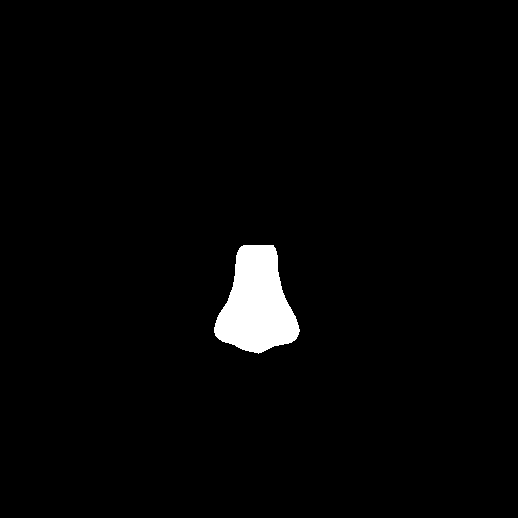}\\

\rotatebox{90}{ Latent diffusion} &
\includegraphics[width=0.133\linewidth]{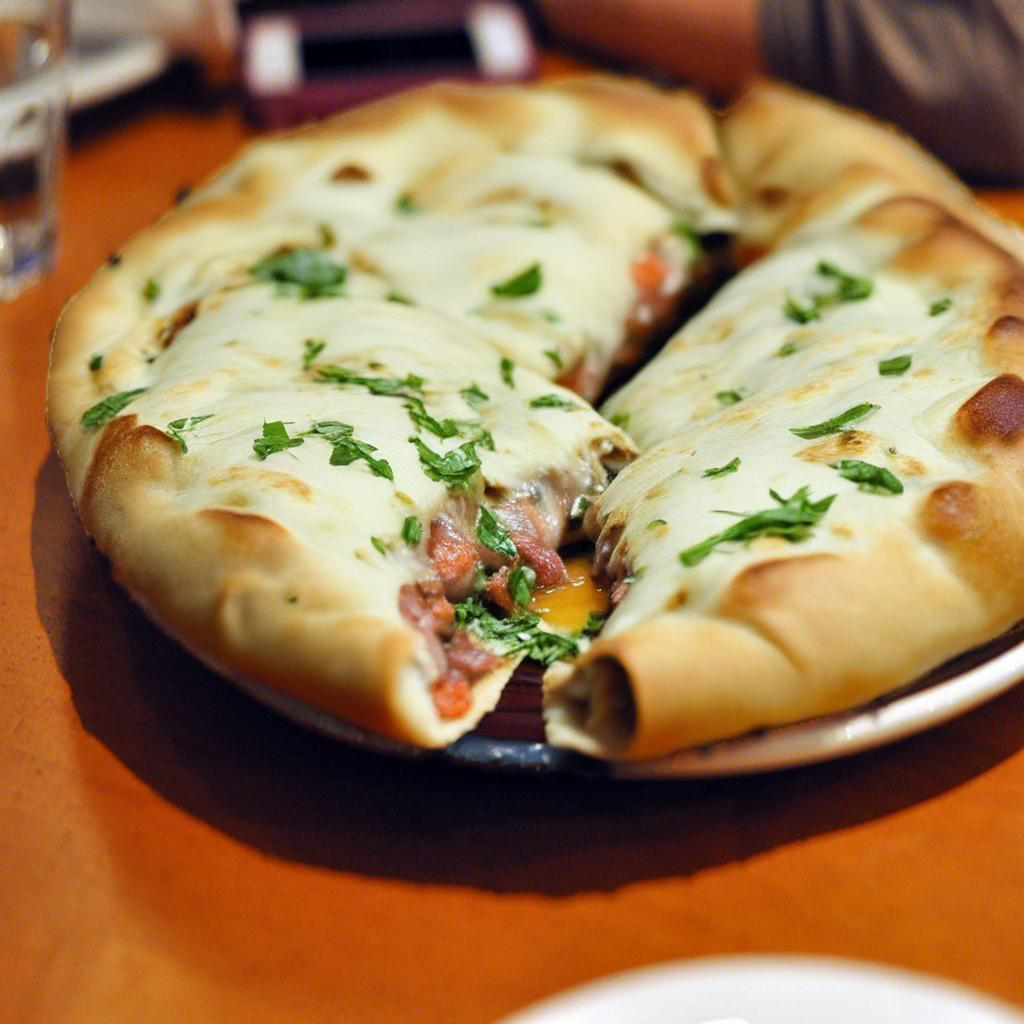} &
\includegraphics[width=0.133\linewidth]{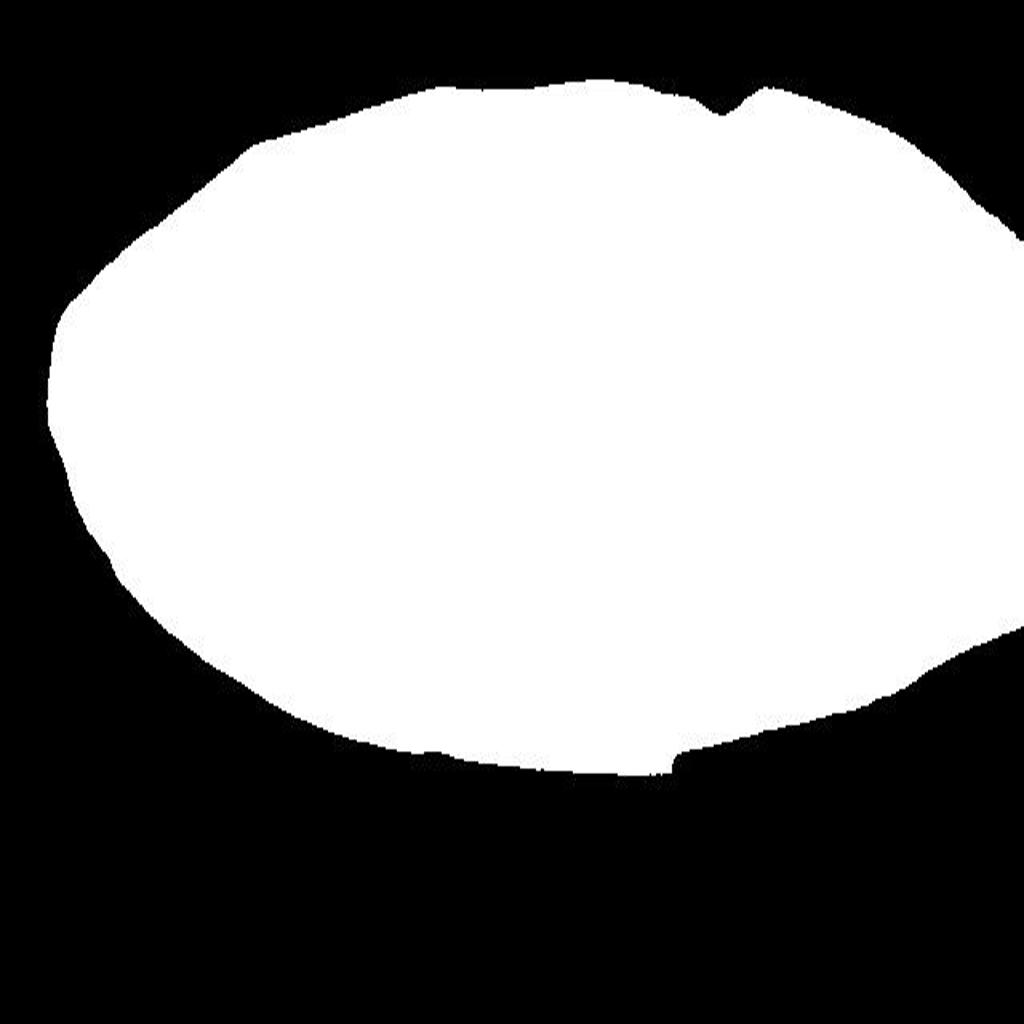} &
\includegraphics[width=0.133\linewidth]{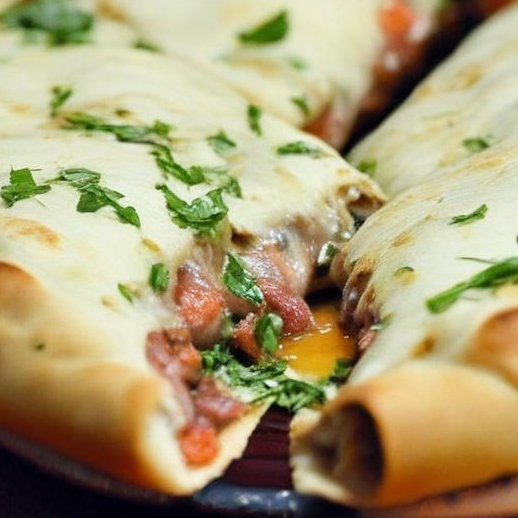} &
\includegraphics[width=0.133\linewidth]{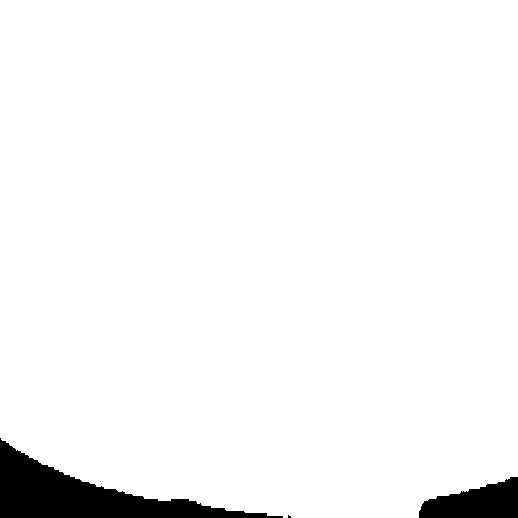} &
\includegraphics[width=0.133\linewidth]{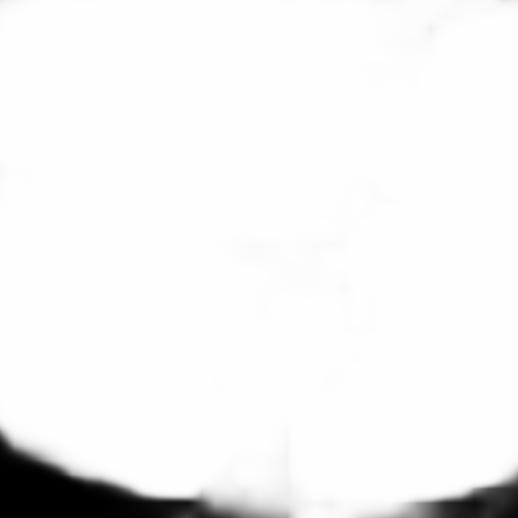}&
\includegraphics[width=0.133\linewidth]{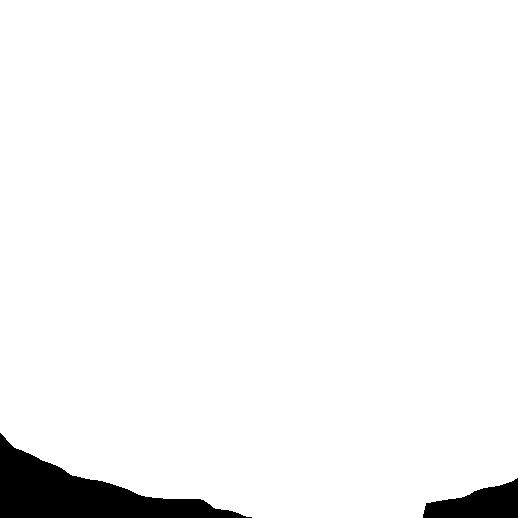}\\

\rotatebox{90}{ $\qquad$ SD-XL} &
\includegraphics[width=0.133\linewidth]{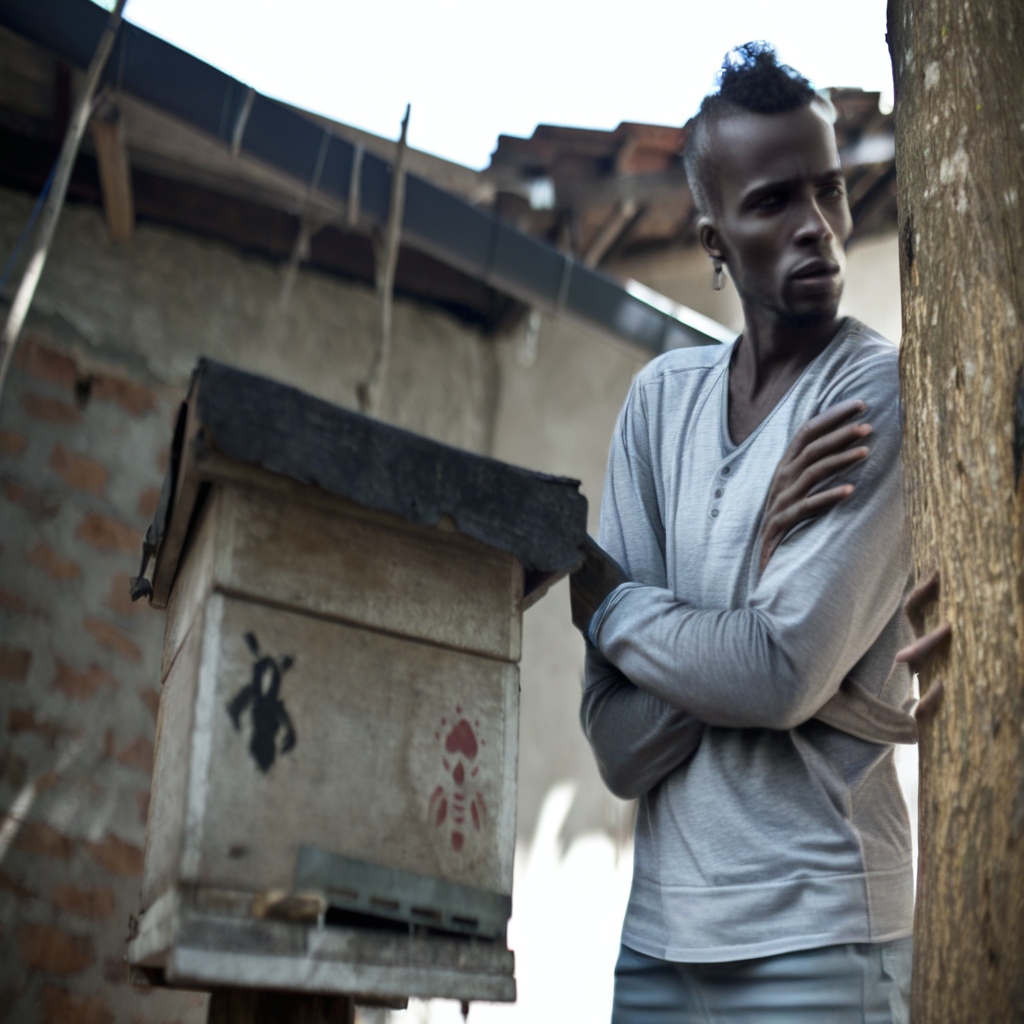} &
\includegraphics[width=0.133\linewidth]{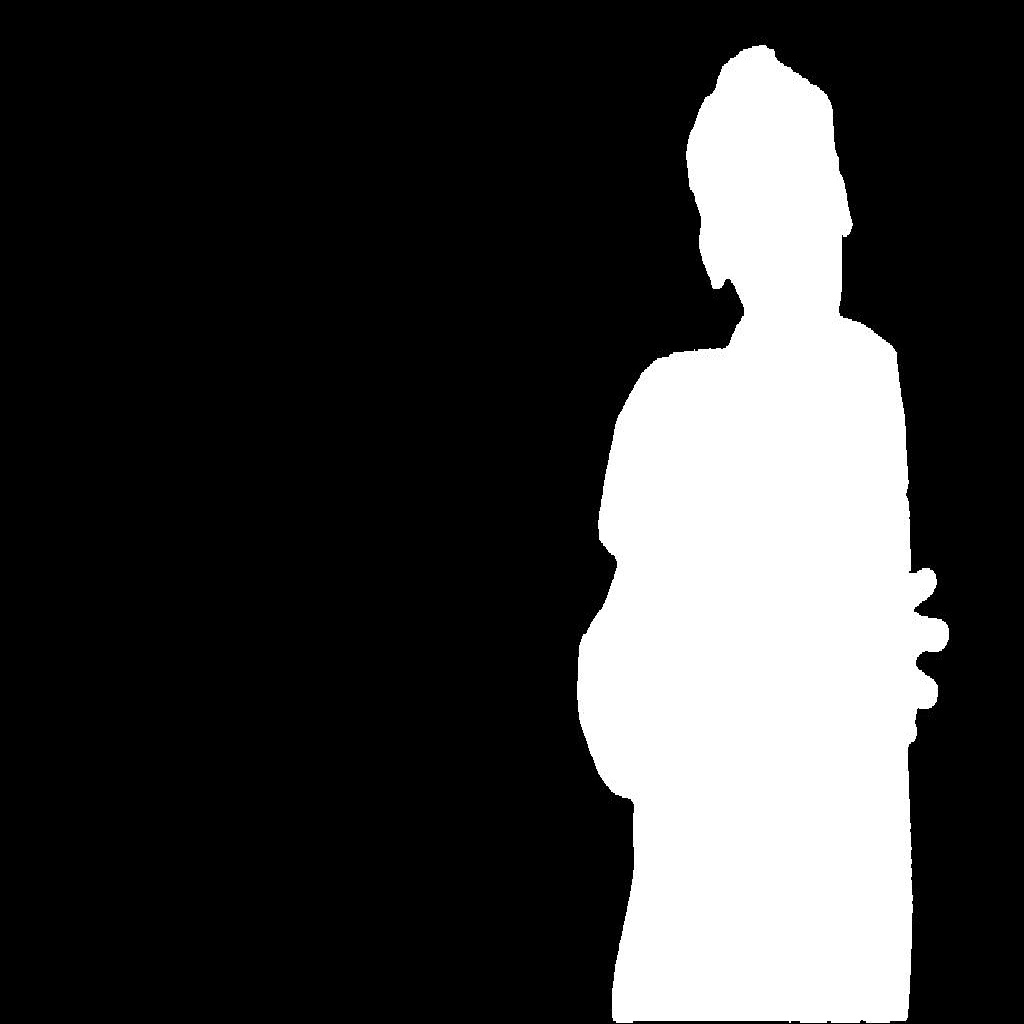} &
\includegraphics[width=0.133\linewidth]{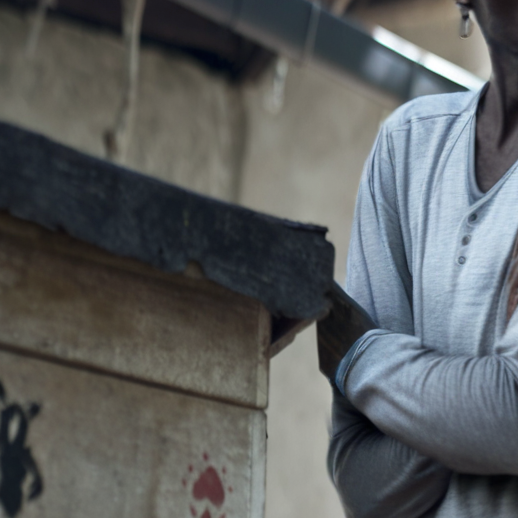} &
\includegraphics[width=0.133\linewidth]{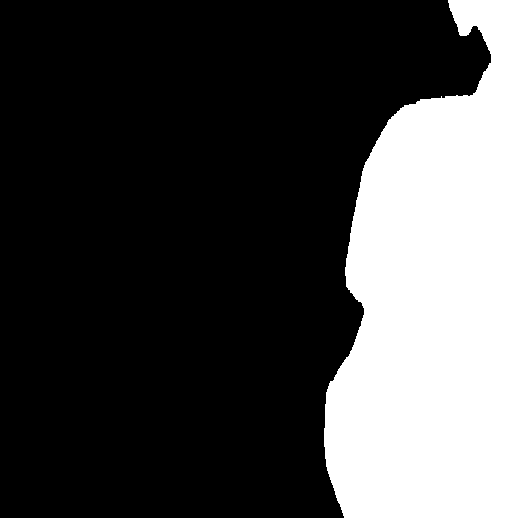} &
\includegraphics[width=0.133\linewidth]{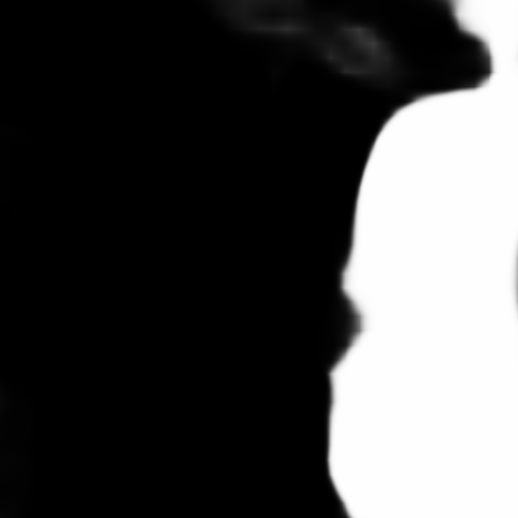}&
\includegraphics[width=0.133\linewidth]{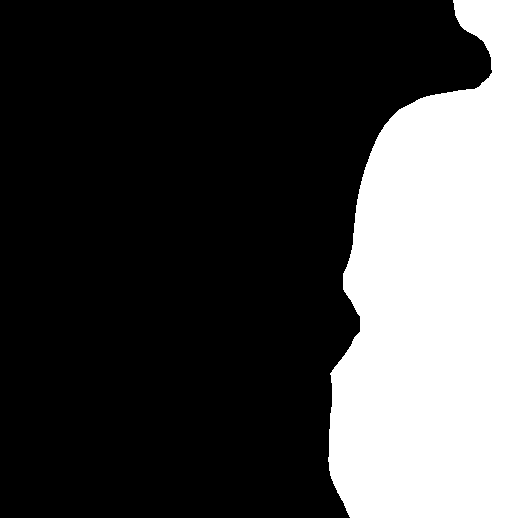}\\

\rotatebox{90}{ $\qquad$ Imagic} &
\includegraphics[width=0.133\linewidth]{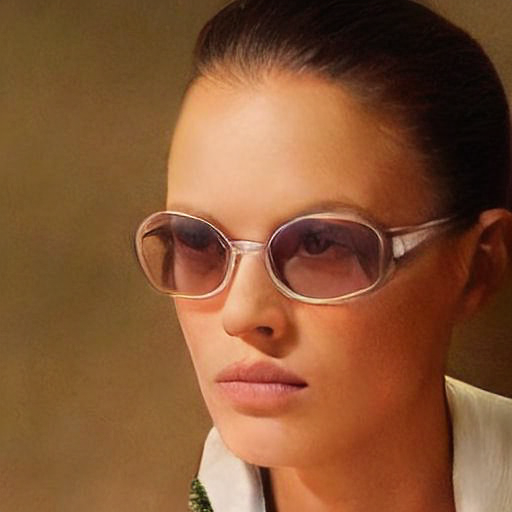} &
\includegraphics[width=0.133\linewidth]{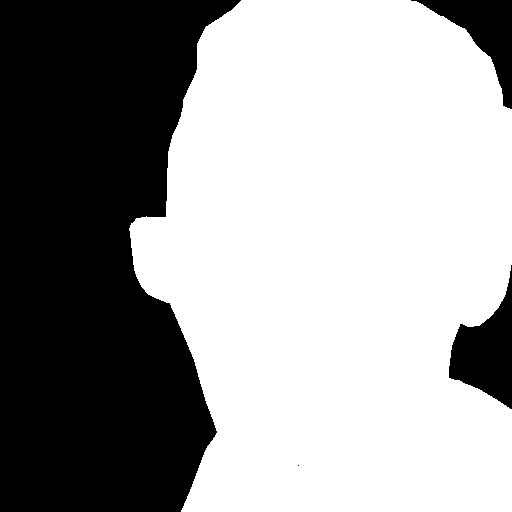} &
\includegraphics[width=0.133\linewidth]{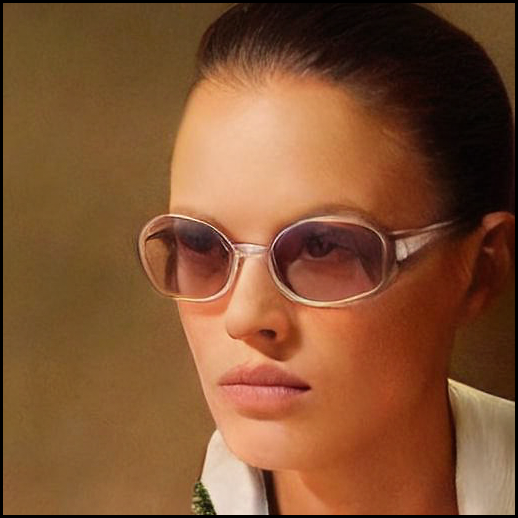} &
\includegraphics[width=0.133\linewidth]{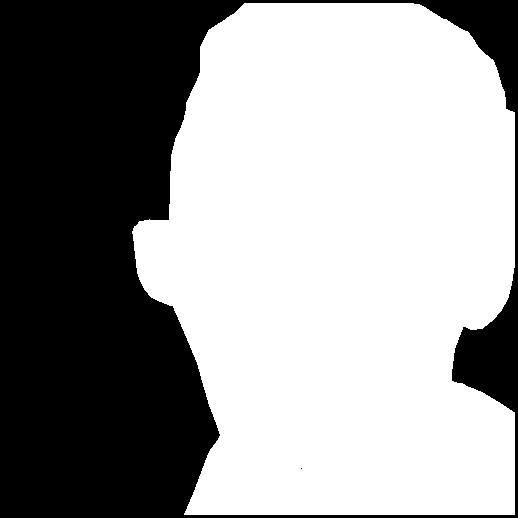} &
\includegraphics[width=0.133\linewidth]{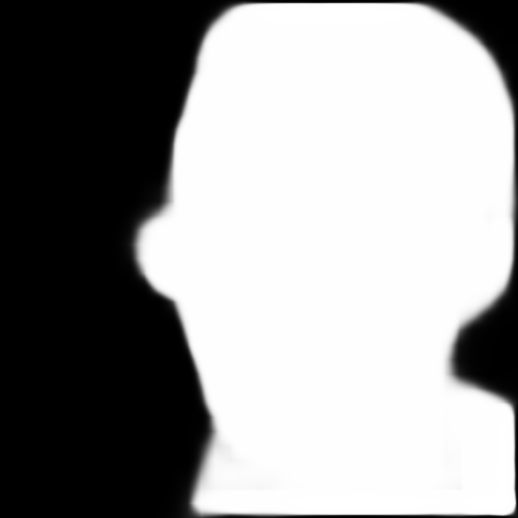}&
\includegraphics[width=0.133\linewidth]{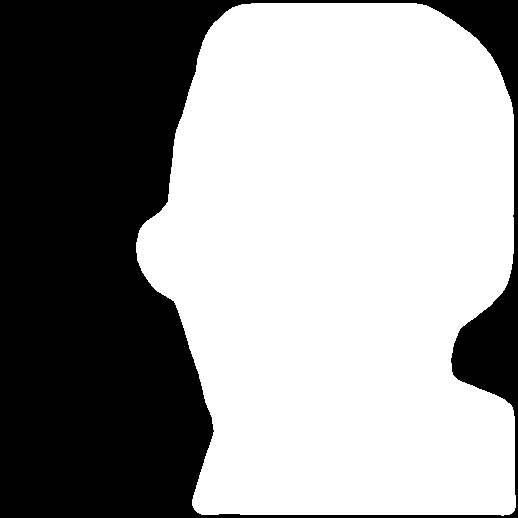}\\

\end{tabular}
\caption{Visual results of the proposed method on tampered images. (a) Original image; (b) ground-truth mask; (c) and (d) corresponding cropped versions; (e) predicted probability map indicating the likelihood of each pixel being manipulated; (f) predicted binary mask after thresholding.}
\Description{Six-column image grid showing tampering localization results for four image generation methods: FLUX.1-dev, Latent diffusion, SD-XL, and Imagic. Columns display the original image, ground-truth mask, cropped image, cropped mask, predicted probability map, and predicted binary mask. The examples include a face, a food image, an outdoor person, and a portrait, with different tampered regions highlighted by the masks. The predicted binary masks closely match the ground-truth masks in all examples.}

\label{fig:good}
\end{figure*}
%\vspace{-1em}
\begin{figure*}[t!]
\centering

\setlength{\tabcolsep}{2pt} % spazio tra colonne

\begin{tabular}{c c c c c c c}
& \textbf{(a)} & \textbf{(b)} & \textbf{(c)} & \textbf{(d)} &\textbf{(e)} & \textbf{(f)} \\

\rotatebox{90}{ $\quad$ FLUX.1-dev} &
\includegraphics[width=0.133\linewidth]{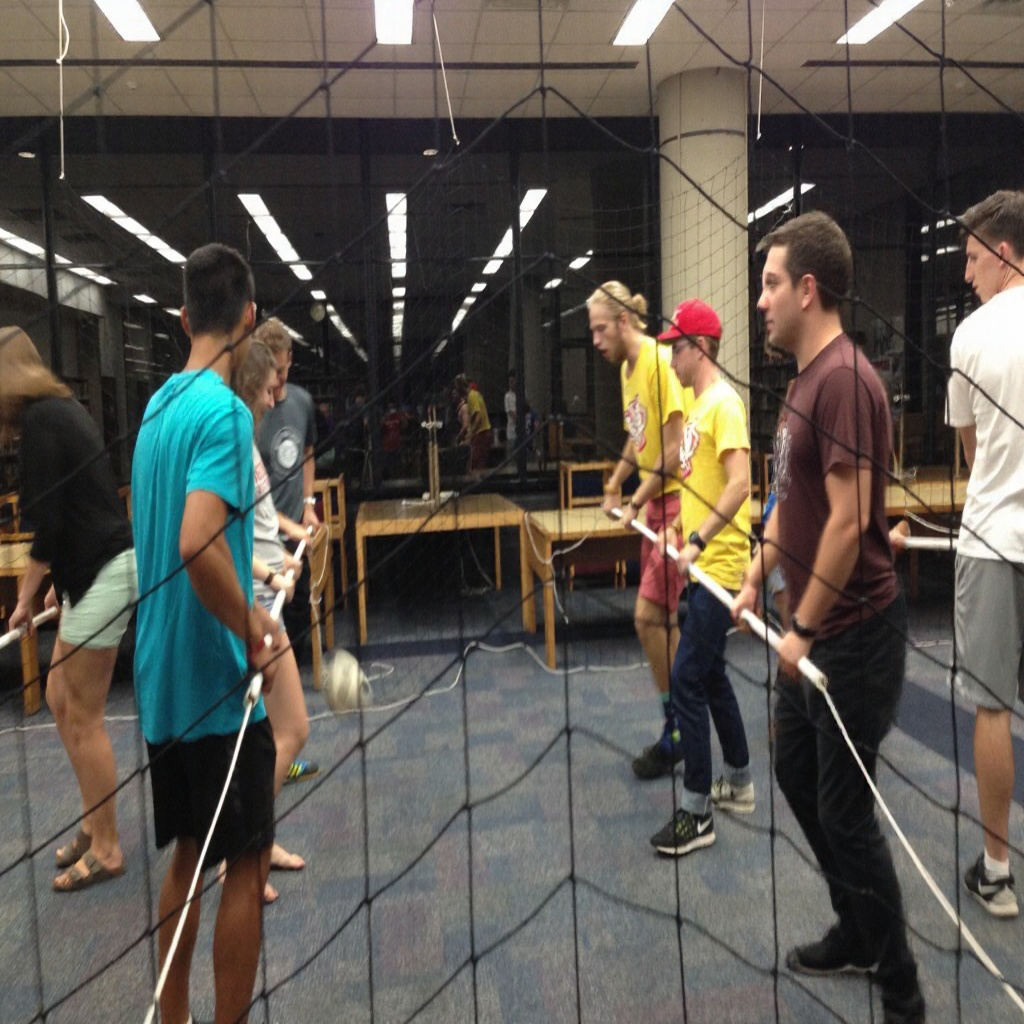} &
\includegraphics[width=0.133\linewidth]{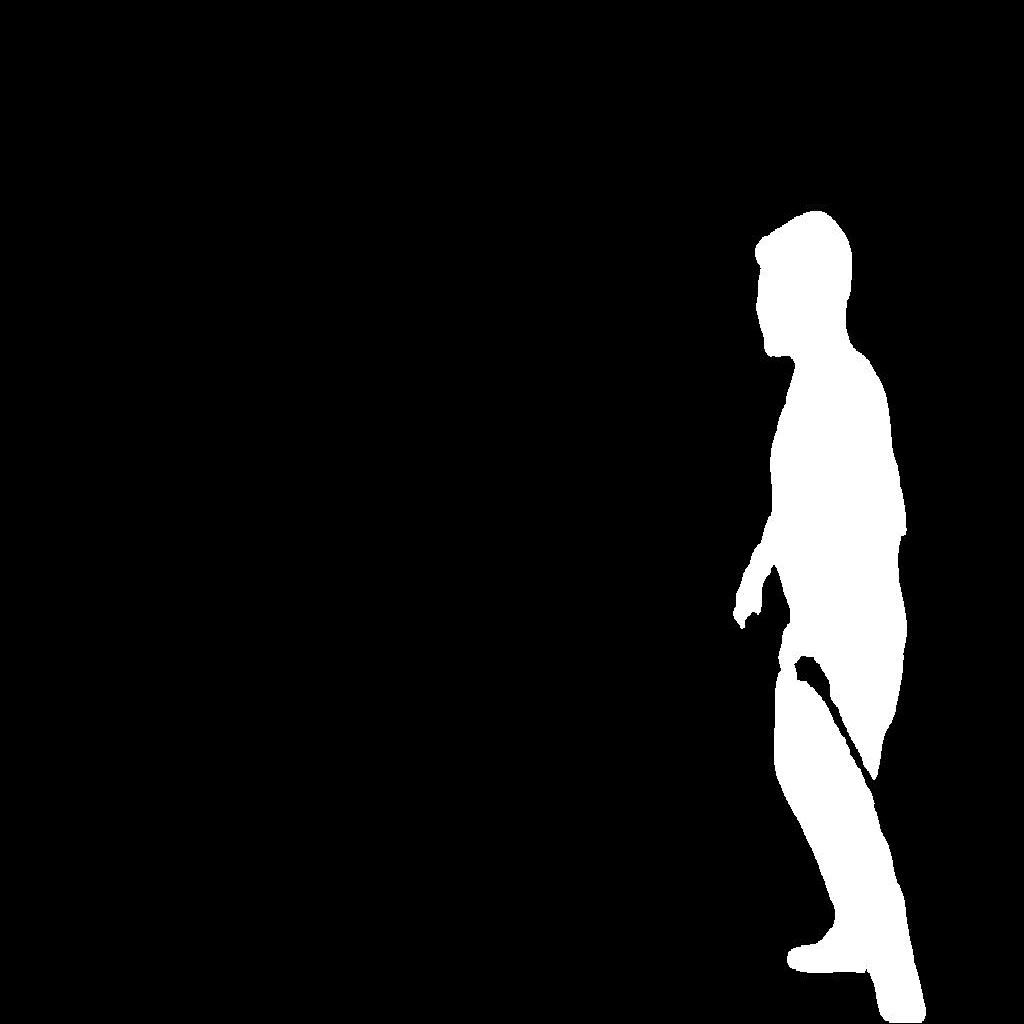} &
\includegraphics[width=0.133\linewidth]{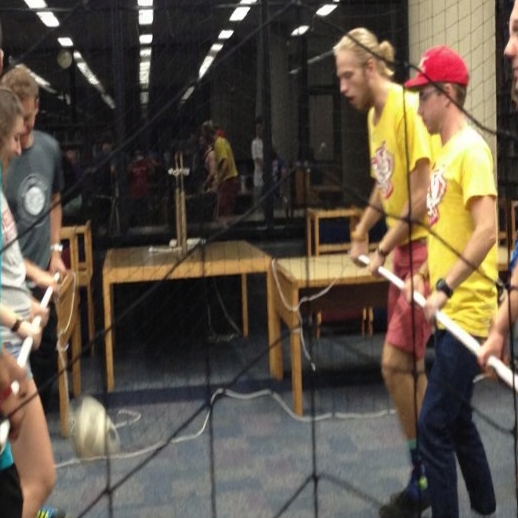} &
\includegraphics[width=0.133\linewidth]{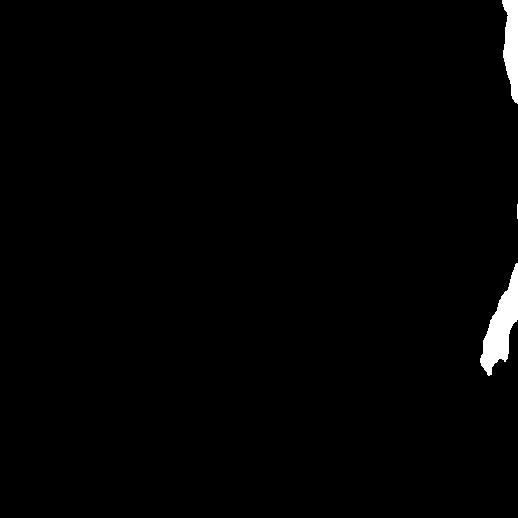} &
\includegraphics[width=0.133\linewidth]{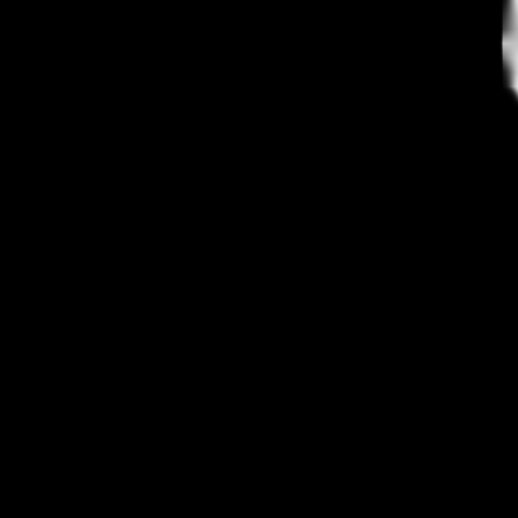}&
\includegraphics[width=0.133\linewidth]{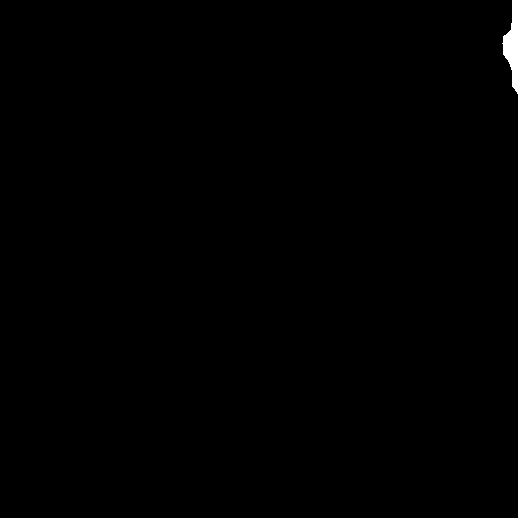}\\

\rotatebox{90}{$\quad$ FLUX.1-dev} &
\includegraphics[width=0.133\linewidth]{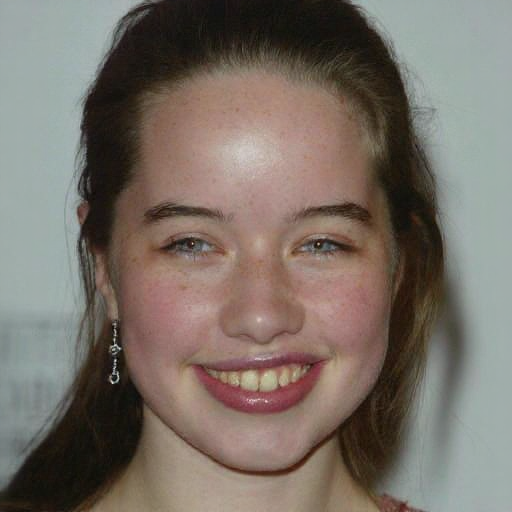} &
\includegraphics[width=0.133\linewidth]{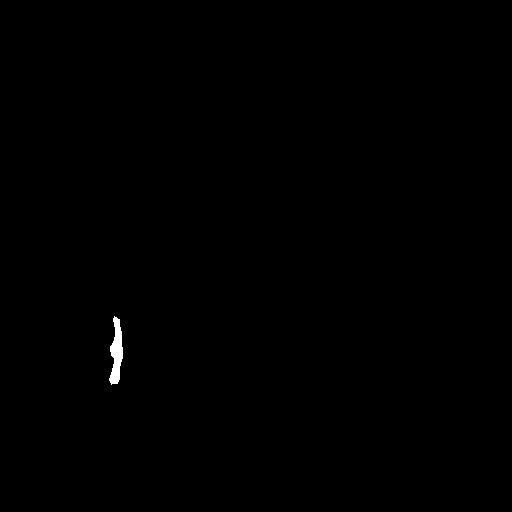} &
\includegraphics[width=0.133\linewidth]{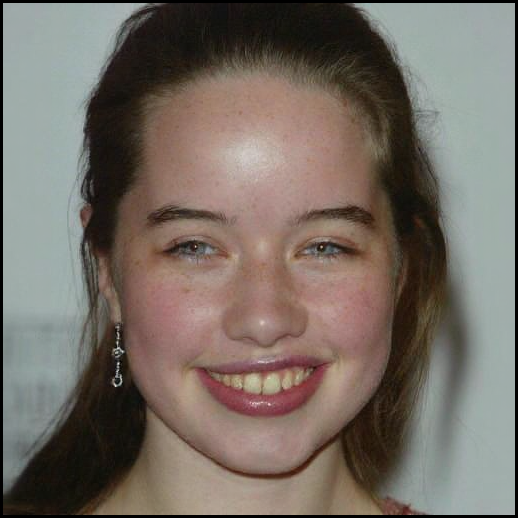} &
\includegraphics[width=0.133\linewidth]{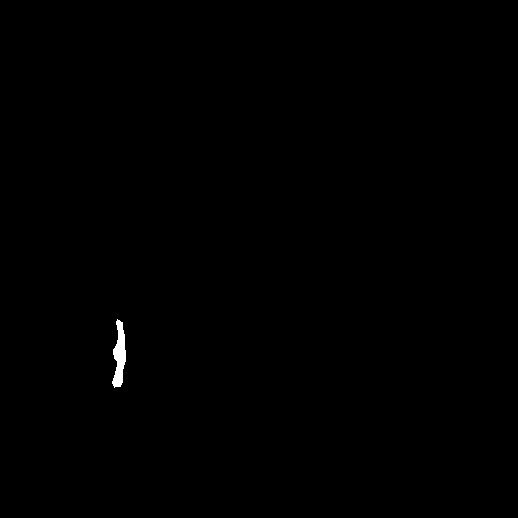} &
\includegraphics[width=0.133\linewidth]{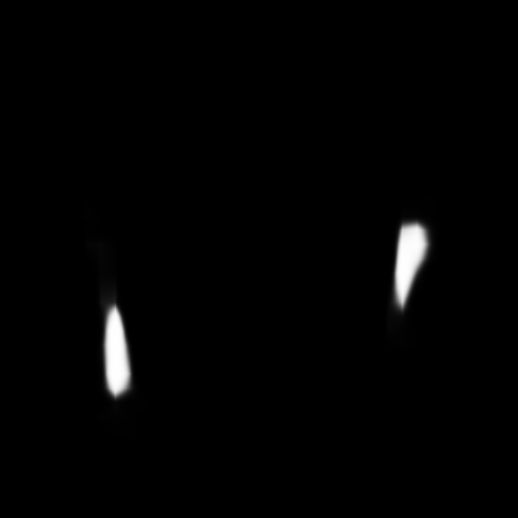}&
\includegraphics[width=0.133\linewidth]{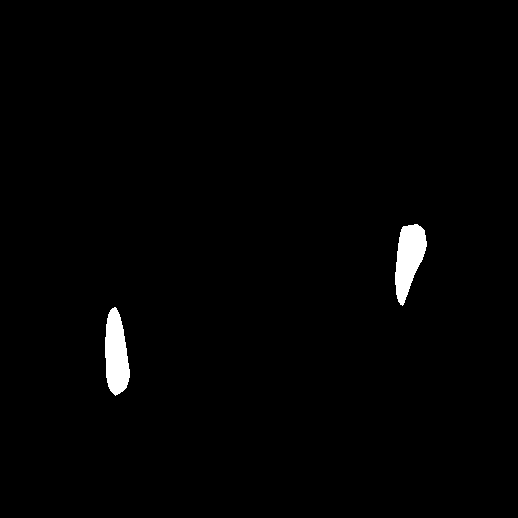}\\

\rotatebox{90}{ $\qquad$ SD-XL} &
\includegraphics[width=0.133\linewidth]{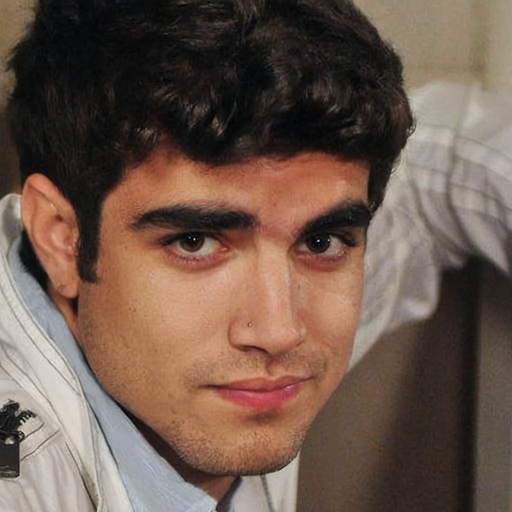} &
\includegraphics[width=0.133\linewidth]{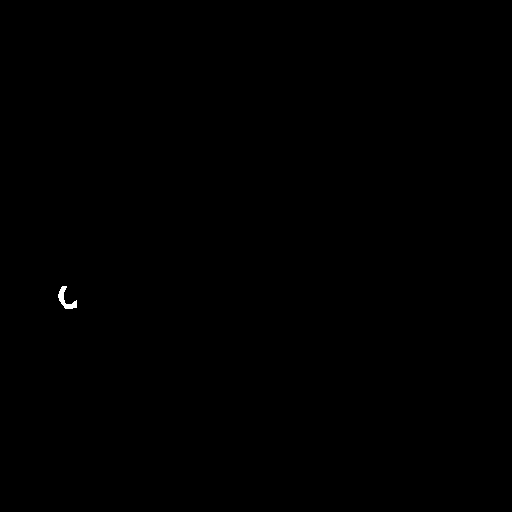} &
\includegraphics[width=0.133\linewidth]{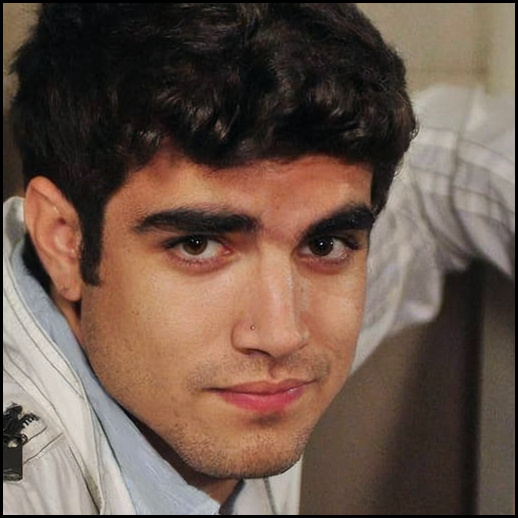} &
\includegraphics[width=0.133\linewidth]{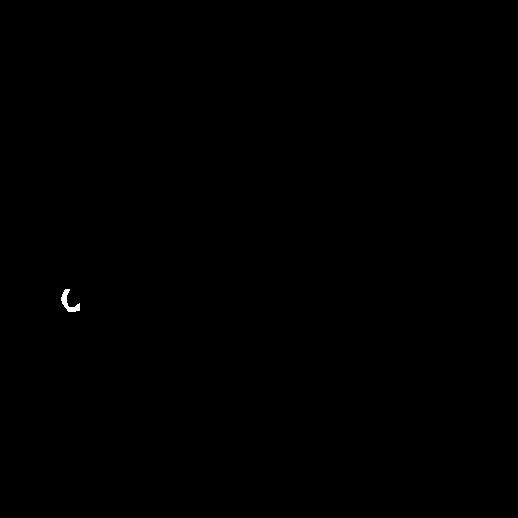} &
\includegraphics[width=0.133\linewidth]{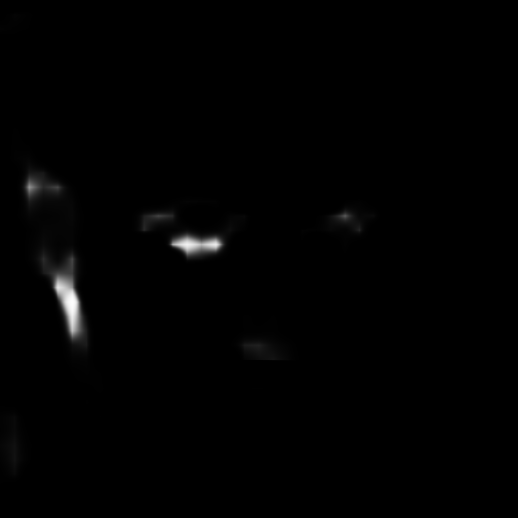}&
\includegraphics[width=0.133\linewidth]{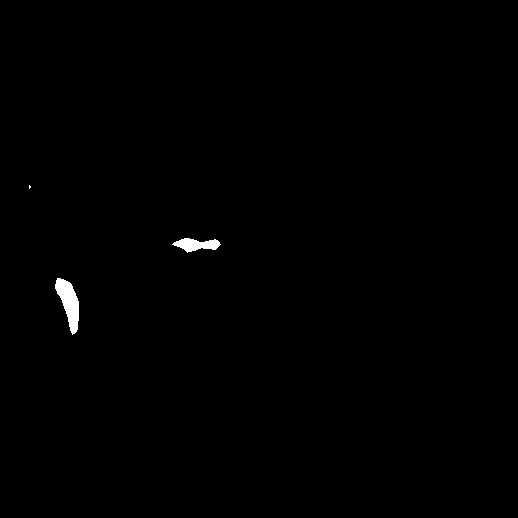}\\

\rotatebox{90}{ Latent diffusion} &
\includegraphics[width=0.133\linewidth]{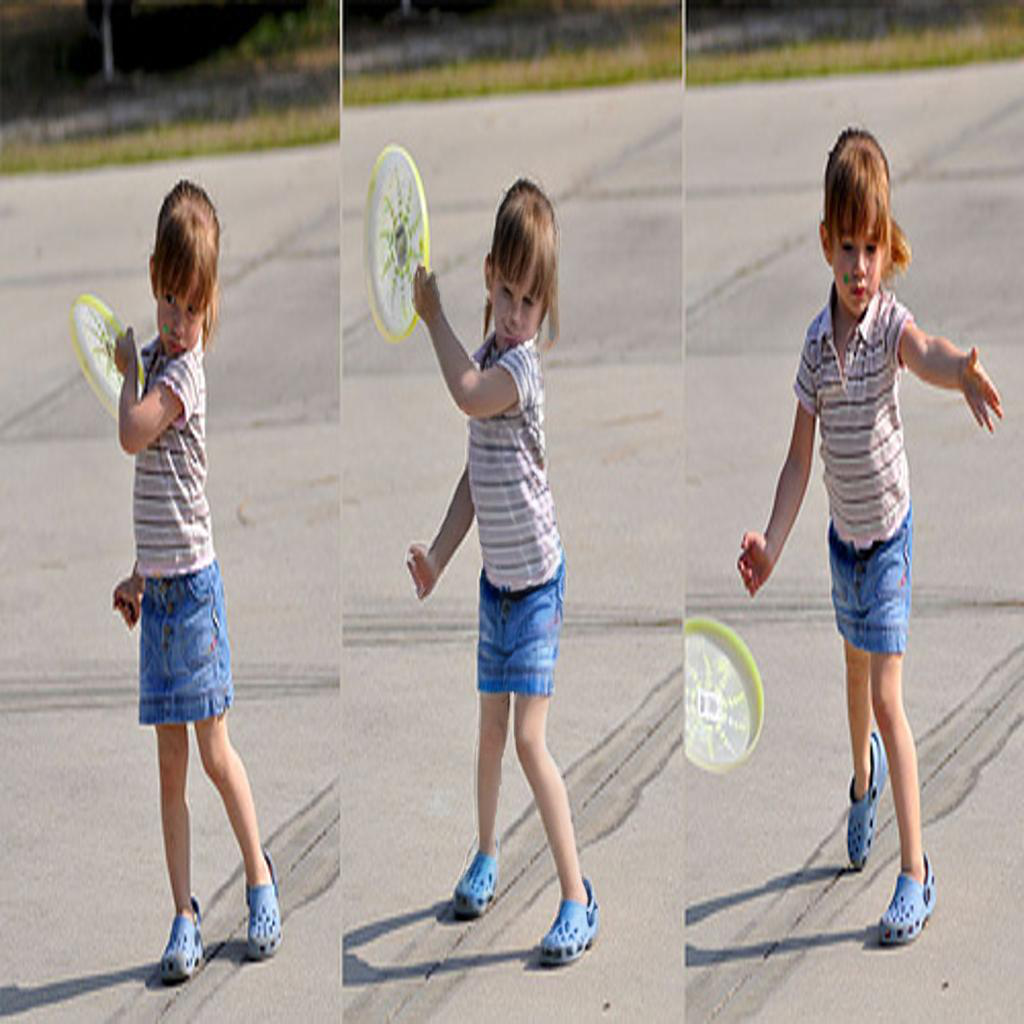} &
\includegraphics[width=0.133\linewidth]{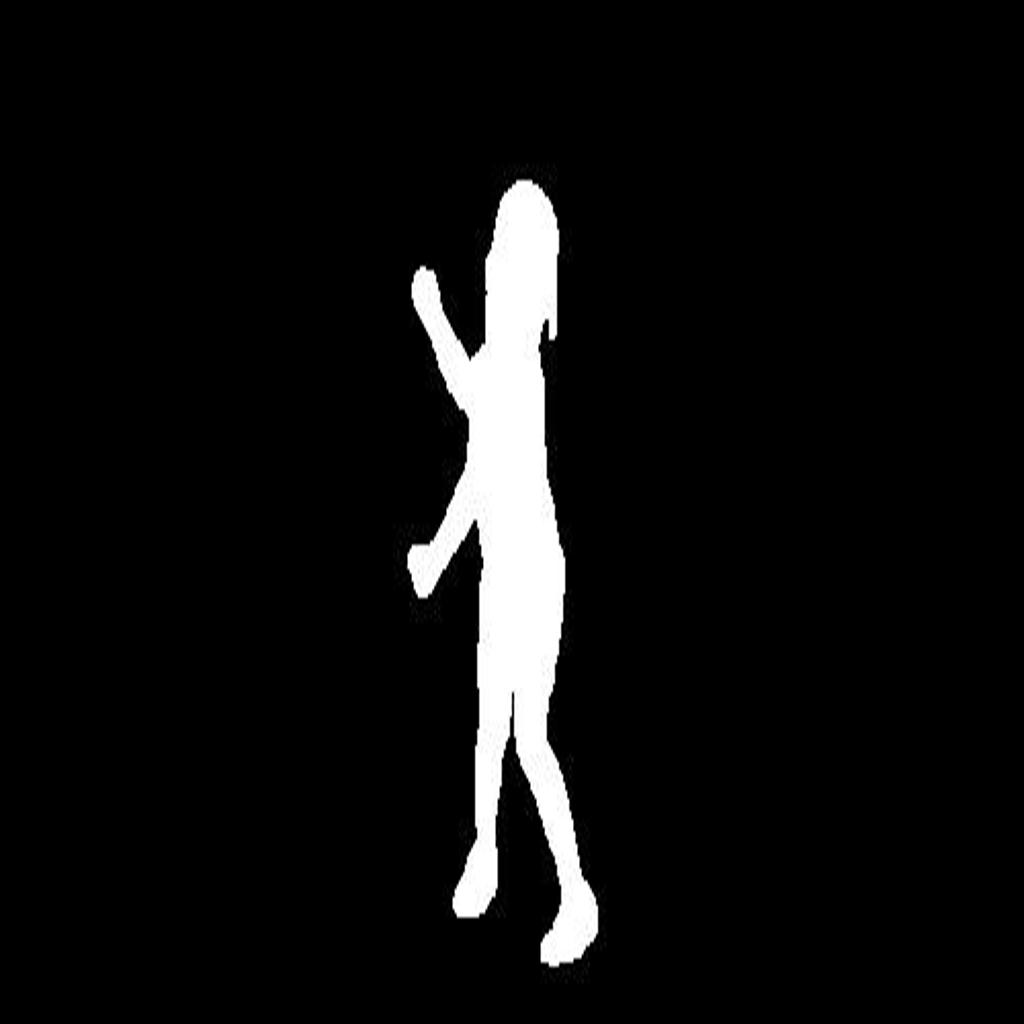} &
\includegraphics[width=0.133\linewidth]{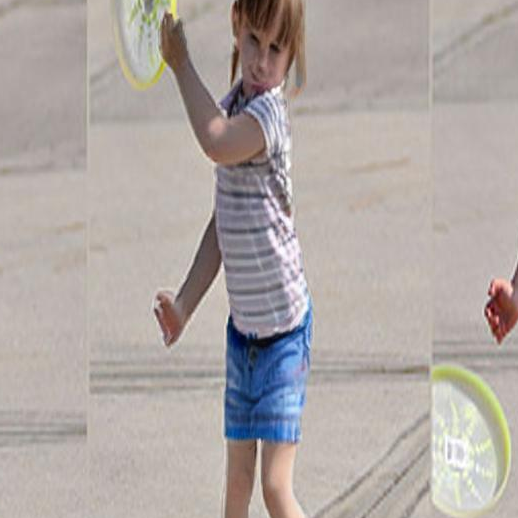} &
\includegraphics[width=0.133\linewidth]{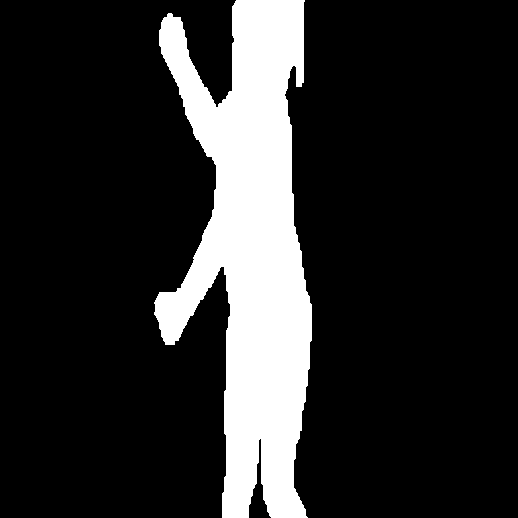} &
\includegraphics[width=0.133\linewidth]{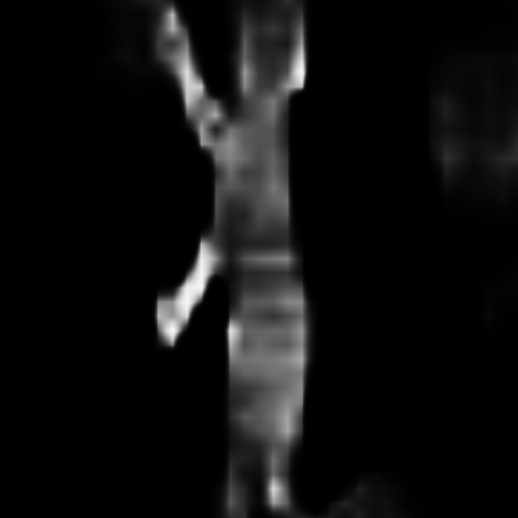}&
\includegraphics[width=0.133\linewidth]{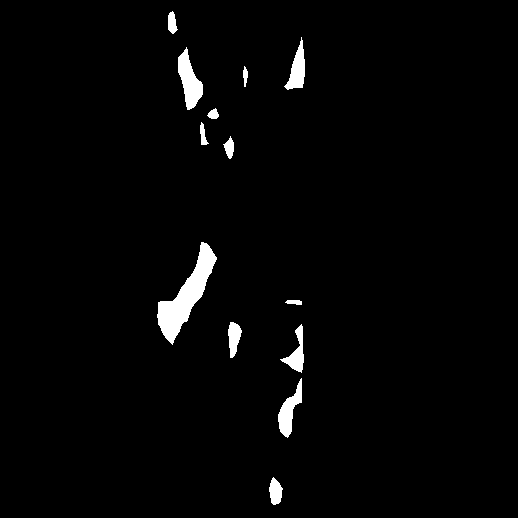}\\

\end{tabular}

\caption{Failure cases of the proposed method on tampered images. (a) Original image; (b) ground-truth mask; (c) and (d) corresponding cropped versions; (e) predicted probability map indicating the likelihood of each pixel being manipulated; (f) predicted binary mask after thresholding.}
\Description{Six-column image grid showing failure cases for tampering localization, with the same layout as Figure 3. Columns display the original image, ground-truth mask, cropped image, cropped mask, predicted probability map, and predicted binary mask. The examples include images generated by FLUX.1-dev, SD-XL, and Latent diffusion. The model struggles with small or thin manipulated regions, producing incomplete reconstructions, over-expanded probability maps, or fragmented low-confidence binary masks compared with the ground-truth annotations.}
\label{fig:bad}
\vspace{-1em}
\end{figure*}

Figure~\ref{fig:good} presents qualitative results that illustrate the effectiveness of the proposed method in localizing tampered regions across a diverse set of image types and generative sources. Each row corresponds to images produced by a different generation model (i.e. FLUX.1-dev~\cite{labs2025flux1kontextflowmatching}, Latent diffusion~\cite{RombachDiffusion}, SD-XL~\cite{podell2024sdxl}, Imagic~\cite{kawar2023imagic}), highlighting the robustness of the proposed approach under varying editing pipelines. It can be observed that the predicted binary masks closely match the ground-truth annotations, accurately capturing the structure and extent of the manipulated regions. The probability maps reveal that the model produces high-confidence responses in tampered areas while maintaining low activation in authentic regions, further confirming its discriminative capabilities.

%Figure~\ref{fig:bad}, in contrast, presents more challenging cases where the proposed method shows some limitations in identifying manipulated regions. From these examples, a few trends can be observed. 
In contrast, Figure~\ref{fig:bad} presents more challenging cases in which the proposed method shows some limitations in identifying manipulated regions. From these examples, several trends can be observed. 
The main issue stems from the aggressive cropping needed to meet the input constraints of DINOv2: when most of the manipulated region falls outside the cropped area, the model may have difficulty detecting the remaining visible portion.
%The main issue is related to the aggressive cropping required to satisfy the input constraints of DINOv2: when the manipulated region lies largely outside the cropped area, the model may struggle to detect the visible portion. 
%Additionally, when the tampered region is very small, the model tends to highlight a slightly larger surrounding area that still overlaps with the ground truth. It is also interesting to notice that the model often attempts to identify symmetric patterns, as can be observed in the third row, where the highlighted region mirrors the manipulated structure despite not corresponding exactly to the ground truth. Finally, the last row shows cases where the manipulated region is approximately localized, but the confidence of the model is relatively low, leading to less accurate final masks.
Additionally, as can be observed in the two middle rows, when the tampered region is very small, the model tends to highlight a slightly larger surrounding area that still partially overlaps with the ground truth.
%It is also interesting to note that the model often attempts to capture symmetric patterns, as illustrated in the third row, where the highlighted region mirrors the manipulated structure despite not exactly matching the ground truth. 
Finally, the last row presents cases in which the manipulated region is approximately localized but the model's confidence remains relatively low, resulting in less accurate final masks.

\subsection{Ablation Studies}
To assess the effectiveness of DINOv2 as backbone, we compare the results obtained extracting features from it with those obtained using CLIP:ViT-L/14~\cite{vitl}.

To address the mismatch between the input resolutions required by the two models, different preprocessing strategies are adopted. For DINOv2, we follow the procedure described in the Preprocessing Section. In contrast, for CLIP, images are first center-cropped to $512 \times 512$ pixels and then resized to $224 \times 224$ pixels.
These distinct strategies are motivated by the characteristics of the training dataset: aggressively cropping high-resolution images to $224 \times 224$ pixels can discard relevant fine-grained details, potentially affecting both detection and localization performance.

Table~\ref{tab:ablation} reports the results for the detection task. In both configurations, the model is trained on the SIDA-Set training split and evaluated on the corresponding test split. The results show a clear improvement when using DINOv2, indicating its stronger capability to capture stable and informative local features, and supporting its selection as the backbone architecture.

\begin{table}[h]
    \centering   

    \begin{tabular}{lcccc}
        \toprule
        Backbone & Real & Fake & Tampered & Overall \\
        \midrule
        CLIP:ViT-L/14 & \underline{95.1} & 95.9 & 69.0 & 86.6 \\
        DINOv2        & 93.1         & \underline{99.8} & \underline{93.3} & \underline{95.4} \\
        \bottomrule
    \end{tabular}
    \caption{Performance comparison with different backbones. Best results are \underline{underlined}.}    
    \label{tab:ablation}
    \vspace{-2em}
\end{table}
\subsection{Model Complexity and Computational Efficiency}
Table~\ref{tab:complexity} compares model complexity and computational efficiency with respect to representative baselines, highlighting the efficiency gains achieved by our approach over \ac{LVLM}-based methods. We include SIDA (7B version)~\cite{ZhenglinSIDA} and the general-purpose \ac{LVLM}s Qwen2.5-VL-7B~\cite{bai2025qwen25} and LISA~\cite{lisa}. So-Fake-R1~\cite{sofake} is not included since its implementation is not available, while methods with significantly lower accuracy than ours are not considered for comparison.

In the comparison, we consider both memory requirements, assessed via total parameter count and model size, and computational efficiency, assessed via inference time. For a fair evaluation, all methods were tested under identical conditions on an NVIDIA L40s GPU, with inference time averaged over 30 randomly sampled images from the SIDA-set (10 per class). To account for initialization effects, a warm-up iteration was performed prior to measurement, and the corresponding inference time was excluded from the reported average. Since the fine-tuned weights on the So-Fake set were not publicly available, we used the original ones. For SIDA~\cite{ZhenglinSIDA}, we followed the prompting strategy described in the original paper, while for \ac{LVLM}s we adopted the prompt reported in~\cite{sofake}.

As shown in Table~\ref{tab:complexity}, our method offers substantial efficiency gains over \ac{LVLM}-based methods. It requires approximately $80\times$ fewer parameters (96.9M vs. 7.71B) and is roughly $40\times$
smaller in model size. It also delivers an approximately $16\times$ speedup in inference, lowering the average per-image processing time from 262.61ms to 16.40ms.  These results demonstrate the significantly lower computational cost of our approach while preserving the ability to perform both detection and localization.
\begin{table}[h]
    \vspace{-0.5em}
    \centering   

    \begin{tabular}{lcccc}
        \toprule
        Method & Parameters $\downarrow$ & Model Size (MB) $\downarrow$ & Time (ms) $\downarrow$\\
        \midrule
        Qwen2.5-VL & 8.29B &  15816 & 1763.58\\
        LISA& 7.70B  &  14727  &652.10\\
        SIDA & 7.71B  & 14744 & 262.61\\
        Ours        &   \underline{96.9M}      & \underline{370} & \underline{16.40} \\
        \bottomrule
    \end{tabular}
    \caption{Comparison of model complexity and computational efficiency across models. Best results are \underline{underlined}.}    
    \label{tab:complexity}
    \vspace{-2em}
\end{table}

\section{CONCLUSIONS}
In this paper, we propose a unified multiclass framework that advances deepfake detection beyond the constraints of traditional binary setting. By addressing the structural bottlenecks of current architectures, we developed an approach that first distinguishes real images from synthetic and tampered content, subsequently activating a lightweight branch for pixel level localization. This architecture ensures that tampered regions are identified with high precision while maintaining computational efficiency. Extensive evaluations on the SID-Set and So-Fake-Set benchmarks demonstrate that our framework outperforms the state-of-the-art in localization, establishing a new baseline for enhanced speed and accuracy in image forensics.
 
While our results demonstrate a significant improvement, achieving universal generalization across an evolving landscape of generative models remains a primary challenge. Future research will first conduct a deeper ablation study, including reimplementations of competing methods, to better assess the contribution of each component. We will then focus on discrepancy learning in the frequency domain to isolate invariant synthetic fingerprints that persist across different generative models, rather than relying on generator-specific artifacts. Targeting such universal signatures is key to establishing robustness in generalized deepfake detection.

\subsubsection*{ACKNOWLEDGEMENTS}
This work was partially supported by the European Union’s Horizon Europe Program under Agreement 101135637 (HEAT Project).
The authors thank the University of Padova, Department of Mathematics ``T. Levi-Civita'', for providing the computational resources used in this work.
\bibliographystyle{unsrt}
\bibliography{main}


\end{document}